\RequirePackage{fix-cm}
\documentclass[twocolumn]{svjour3}          %
\smartqed  %
\usepackage{graphicx}
\usepackage{times}
\usepackage{epsfig}
\usepackage{amsmath}
\usepackage{amssymb}
\usepackage{amsmath} %
\usepackage{color}
\usepackage{url}
\usepackage{subfigure}
\usepackage{multirow,bm}
\usepackage{wrapfig}
\usepackage[authoryear]{natbib}
\bibpunct[; ]{(}{)}{,}{a}{}{;}

\begin{document}

\title{Subjects and Their Objects: Localizing Interactees for a\\Person-Centric View of Importance}%

\author{Chao-Yeh Chen         \and
        Kristen Grauman %
}

\institute{C.-Y. Chen \at
              Department of Computer Science, \\
              University of Texas at Austin, Austin, TX, USA \\
              \email{chaoyeh@cs.utexas.edu}           %
           \and
           K. Grauman \at
              Department of Computer Science, \\
              University of Texas at Austin, Austin, TX, USA \\
              \email{grauman@cs.utexas.edu}
}

\maketitle

\makeatletter{}%
\begin{abstract}
Understanding images with people often entails understanding their \emph{interactions} with other objects or people.  As such, given a novel image, a vision system ought to infer which other objects/people play an important role in a given person's activity.  However,
 existing methods are limited to learning action-specific interactions (e.g., how the pose of a tennis player relates to the position of his racquet when serving the ball) for improved recognition, making them unequipped to reason about novel interactions with actions or objects unobserved in the training data.

We propose to predict the ``interactee" in novel images---that is, to localize the \emph{object} of a person's action.  Given an arbitrary image with a detected person, the goal is to produce a saliency map indicating the most likely positions and scales where that person's interactee would be found.  To that end, we explore ways to learn the generic, action-independent connections between (a) representations of a person's pose, gaze, and scene cues and (b) the interactee object's position and scale.
We provide results on a newly collected UT Interactee dataset spanning more than 10,000 images from SUN, PASCAL, and COCO.  We show that the proposed interaction-informed saliency metric has practical utility for four tasks:  contextual object detection, image retargeting, predicting object importance, and data-driven natural language scene description.  All four scenarios reveal the value in linking the subject to its object in order to understand the story of an image.

\end{abstract}
\makeatletter{}%
\section{Introduction}\label{sec:intro}

\begin{figure}[t]
\centering
\includegraphics[width=0.5\textwidth]{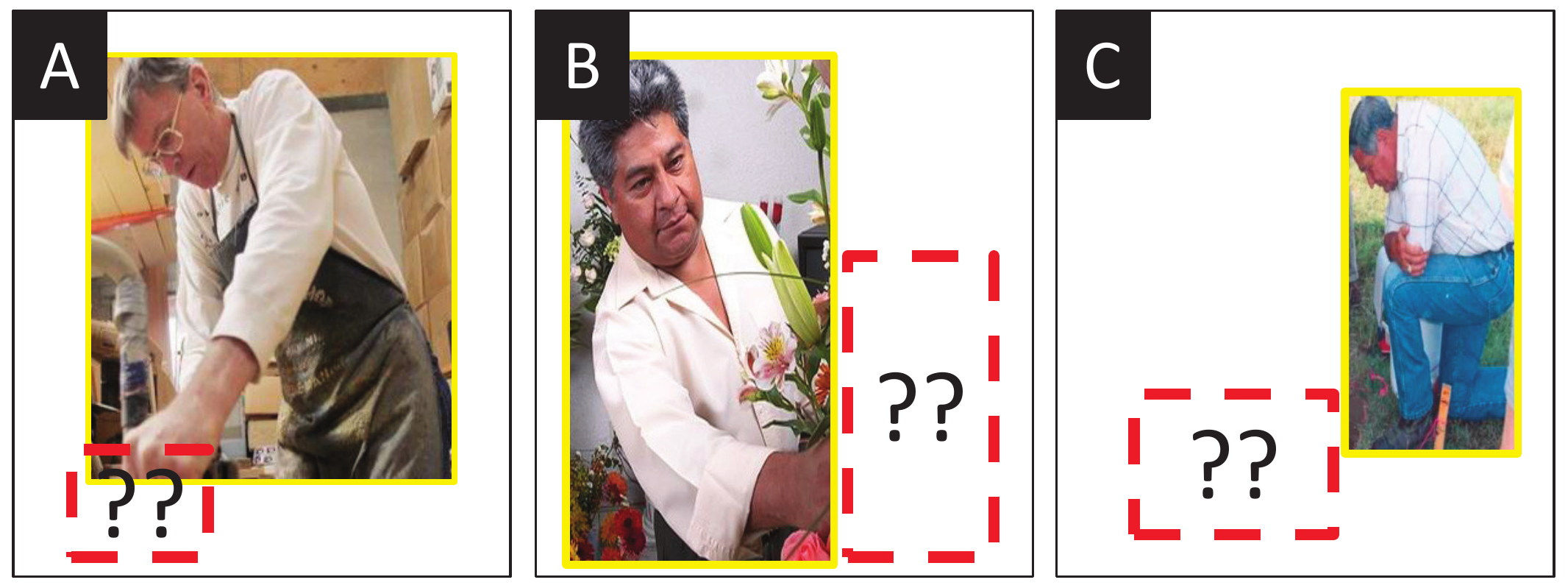}
   \caption{Despite the fact we have hidden the remainder of the scene, can you infer where is the object with which each person is interacting?  Our goal is to predict the position and size of such ``interactee" objects in a \emph{category-independent} manner, without assuming  prior knowledge of the either the specific action or object types.}\label{fig:concept}
\end{figure}

\begin{figure*}[t]
\centering
\includegraphics[width=0.8\textwidth]{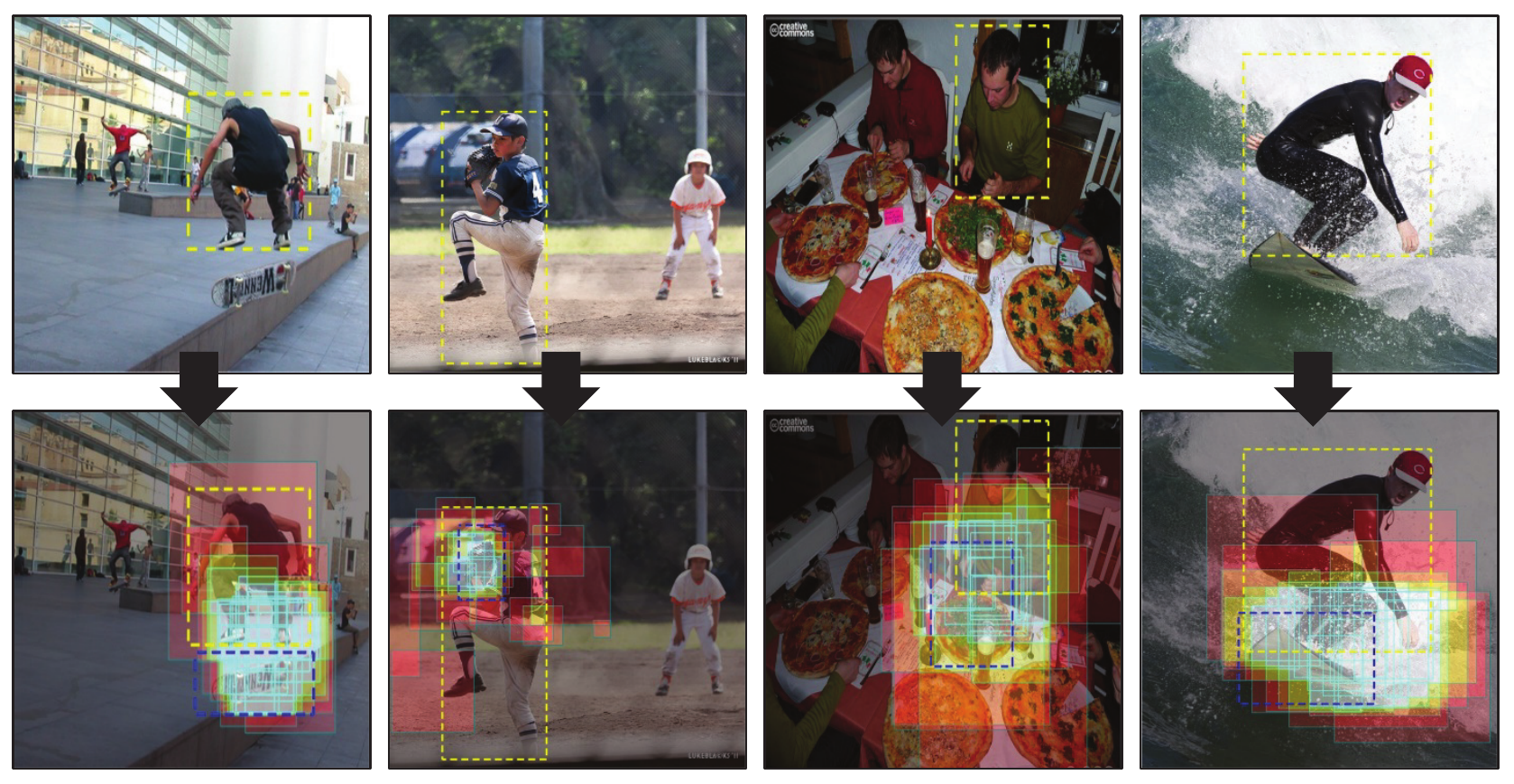}
\caption[Example interactee localizations]{Example test instances processed by our method.  Given an image with a detected person (top row), our method returns an \emph{interaction-informed} saliency map (bottom row).  This map captures the positions and scales of regions where the system expects to see an interactee.  Its predictions are the result of learning action- and object-independent cues that indicate ``where to look" for the interactee relative to the person.  In these successful cases, even though the system posseses no object detector for the highlighted objects, it can nonetheless isolate those regions of interest based on learned cues about where the person is looking, his/her body pose, and the general scene semantics. For example, our system infers the interactee's position and scale across different types of interactions, such as holding, riding, and eating. See Sec.~\ref{sec:experiment} for details and more examples, including failure cases.}\label{fig:intro-examples}
\end{figure*}

Understanding human activity is a central goal of computer vision with a long history of research.  Whereas earlier work focused on precise body pose estimation and analyzed human gestures independent of their surroundings, recent research shows the value in modeling activity in the context of \emph{interactions}.   An interaction may involve the person and an object, the scene, or another person(s).  For example, a person \emph{reading} reads a book or paper; a person \emph{discussing} chats with other people nearby; a person \emph{eating} uses utensils to eat food from a plate.  In any such case, the person and the ``interactee" object (i.e., the book, other people, food and utensils, etc.) are closely intertwined; together they define the story portrayed in the image or video.

Increasingly, research in human action recognition aims to exploit this close connection~\citep{peursum-iccv2005,gupta-pami2009,desai-2010,Yao:2010:context-of-object-and-human,Bangpeng:2010:grouplet,sclaroff-eccv2010,Ferrari:2012:weakly_supervised_learning_interactions,delaitre-eccv2012}.  In such methods, the goal is to improve recognition by leveraging human action in concert with the object being manipulated by the person.  However, prior work is restricted to a closed-world set of objects and actions, and assumes that during training it is possible to learn patterns between a particular action and the particular object category it involves.  For example, given training examples of \emph{using a computer}, typical poses for typing can help detect the nearby computer, and vice versa; however, in existing methods, this pattern would not generalize to help make predictions about, say, a person operating a cash register.  Furthermore, existing work largely assumes that the interactions of interest involve a direct manipulation, meaning that physical contact occurs between the person and the interactee.

We seek to relax these assumptions in order to make predictions about novel, unseen human-object interactions.  In particular, we consider the following question: \emph{Given a person in a novel image, can we predict the location of that person's ``interactee"---the object or person with which he interacts---even without knowing the particular action being performed or the category of the interactee itself?}  In terms of English grammar, the interactee is the \emph{direct object}, i.e., the noun that receives the action of a transitive verb or shows the results of the action.  Critically, by posing the question in this manner, our solution cannot simply exploit learned action-specific poses and objects.  Instead, we aim to handle the open-world setting and learn generic patterns about human-object interactions.  In addition, we widen the traditional definition of an interactee to include not only directly manipulated objects, but also untouched objects that are nonetheless central to the interaction (e.g., the poster on the wall the person is reading).

Our goal is challenging.  %
  A naive approach might simply prioritize objects based on their proximity to a person.  However, as we will show in results, doing so is inadequate---not only because many objects of various scales may closely surround a person, but also because the person need not be touching the interactee (e.g., a supermodel looks in a mirror).  Furthermore, the task is not equivalent to detecting an action, since an action may have multiple interchangeable objects with disparate localization parameters (e.g., planting a flower vs.~planting a tree)

\begin{figure*}[t]
\centering
\includegraphics[width=\textwidth]{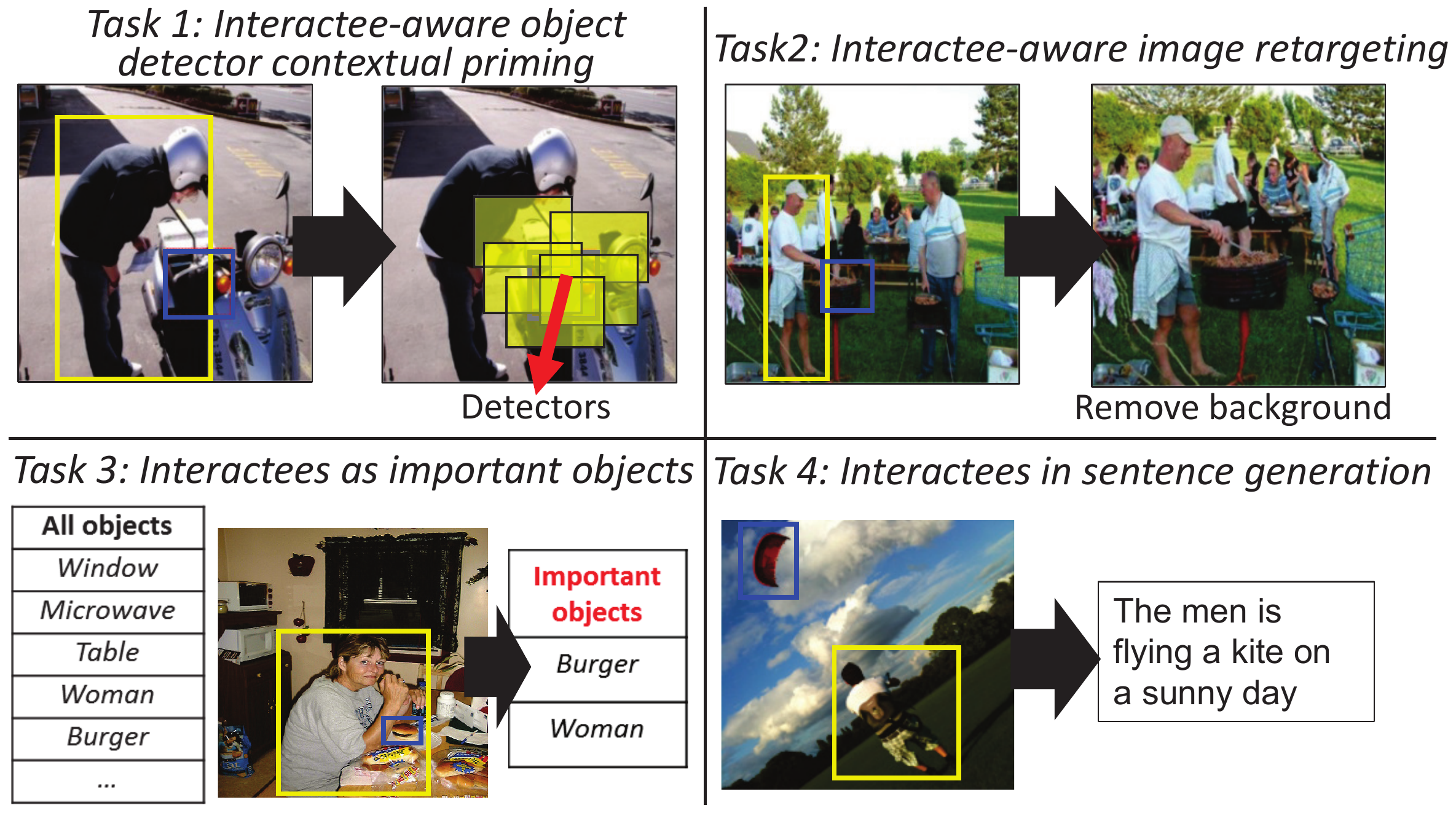}
   \caption{Once we learn to predict where an interactee is likely to be found, we can tackle four fine-grained human-object interaction tasks: (1) \textbf{object detection} exploiting a contextual prior given by the interactee's likely placement, (2) \textbf{image retargeting} where the preference is to preserve not only the subject (person), but also the key object of his interaction, (3) \textbf{object importance} ranking, where we prioritize mentioning an object that is involved in the interaction, and (4) \textbf{sentence descriptions} of images, where we learn to focus a short natural language description of the scene to include the major players (both subject and object).}\label{fig:taskoverview}
\end{figure*}

Why, then, should our goal be possible?  Are there properties of interactions that transcend the specific interactee's category and can be learned?  Figure~\ref{fig:concept} suggests that, at least for human observers, it is plausible.   In these examples, without observing the interactee object or knowing its type, one can still infer the interactee's approximate position and size.  For example, in image A, we may guess the person is interacting with a small object in the bottom left.
We can do so because we have a model of certain pose, gaze, and scene layout patterns that exist when people interact with a person/object in a similar relative position and size.   We stress that this is without knowing the category of the object, and even without (necessarily) having seen the particular action being performed.

Based on this intuition, our idea is to learn from data how the properties of a person in an image (the ``subject'') relate to the interactee localization parameters (the ``object'').  Given instances labeled with both the person and interactee outlines---from a variety of activities and objects---we train a model that can map observed features of the person and scene to a distribution over the interactee's position and scale.  Then, at test time, given a novel image and a detected person, we predict the most likely places the interactee will be found.

In particular, we explore both learned and hand-crafted descriptors for the task.  The learned descriptors use convolutional neural networks (CNNs) pre-trained for object recognition, then fine-tuned for the interactee localization tasks.  Using content from both the person region and the entire scene, these descriptors are free to learn any signal about the interaction and scene layout that is predictive of finding the region with the interactee.  On the other hand, the hand-crafted descriptors more explicitly encode cues likely to inform about interactees' placement, such as the subject's eye gaze, body pose, and relative positioning in the scene.

To build a predictive model from these features, we develop and compare two methods---one non-parametric, one parametric---that can accommodate varying levels of annotated training data.  Our non-parametric approach retrieves annotated examples similar to a test instance in terms of their interactions, then computes a locally weighted regression model to infer the new localization map.  Our alternative parametric approach uses Mixture Density Networks (MDNs)~\citep{Bishop:1994:mixturedensity} to generate a mixture model over scales and positions on the fly, where the connection between the image features and mixture model parameters are directly learned.

Our method can be seen as an \emph{interaction-informed} metric for object saliency: it highlights regions of the novel image most likely to contain objects that play an important role in summarizing the image's content.  See Figure~\ref{fig:intro-examples}.

With this in mind, we develop four novel tasks to leverage the interactee predictions, as previewed in Figure~\ref{fig:taskoverview}.  First, we consider how they can improve the accuracy or speed of an existing object detection framework, by guiding the detector to focus on areas that are involved in the interaction.   Second, we use interactees  to assist in image retargeting. In this task, the image is automatically resized to a target aspect ratio by removing unimportant content and preserving the parts related to the person and  his interactee.

In the third and fourth tasks, we examine the interactee as a means to answer the ``what to mention" question about image description.  As object recognition techniques gain more solid footing, it is now valuable to formulate the task not merely in terms of labeling every object, but also in terms of identifying which, among all true detections, a human observer would bother to include when describing the scene.  To this end, we explore using interactees to rank order objects in the scene by their importance, i.e., their probability of being mentioned~\citep{spain-eccv2008,spain-ijcv2011,berg-importance}.  Then, as the fourth and final task, we map the image to a natural language sentence description, using a retrieval-based approach that finds images with similar interactions in a captioned database.

To facilitate our data-driven strategy as well as quantitative evaluation, we collect interaction annotations for over 15,000 images containing people from the Common Objects In Context (COCO) dataset~\citep{Lin:2014:coco}, and hundreds of images containing people from PASCAL Actions~\citep{Everingham:2010:pascal-voc} and SUN scenes~\citep{Xiao:2010:SUN}.  Results on these challenging datasets evaluate our methods' localization accuracy compared to several baselines, including an existing high-level ``objectness" saliency method~\citep{Ferrari:2010:objectness} and a naive approach that simply looks for interactees nearby a person.  We also perform a human subject study to establish the limits of human perception for estimating unseen interactees.  Finally, we demonstrate practical impact for the four tasks discussed above.

To recap, our main contributions are:
\begin{itemize}
\item We define the problem of interactee localization, which to our knowledge has not been considered in any prior work.\footnote{This manuscript is an extension of our initial conference paper~\citep{chaoyeh-accv2014}.  Please see the cover letter for an overview of the new additions to both the method and results that are contained in this article.}

\item We develop learned person and scene embeddings that preserve information relevant to human-object interactions.  We further introduce and analyze two learning frameworks for making interactee predictions.

\item We gather a large annotated dataset for human-object interactions spanning hundreds of diverse action types, which is necessary to train and evaluate methods for the proposed task.  It will be made publicly available.

\item We analyze the performance of our approaches compared to multiple baselines and existing methods, including methods from the object importance and object proposal literature.  We also examine the impact of learned versus hand-crafted features for our task.

\item We demonstrate the practical value of interactees for four tasks: object detection, image retargeting, object importance, and image description.
\end{itemize}

In the following, we next review related work (Sec.~\ref{sec:related}).  Then we introduce our approach (Sec.~\ref{sec:algorithm}), including systematically defining what qualifies as an interactee, explaining our data collection effort, defining our learning models, and developing the four tasks to exploit it.  Finally, we provide experimental results in Sec.~\ref{sec:experiment}.

\makeatletter{}%
\section{Related Work}\label{sec:related}

\paragraph{Human-object interactions for recognition}

A great deal of recent work in human activity recognition aims to jointly model the human and the objects with which he or she interacts~\citep{peursum-iccv2005,gupta-pami2009,desai-2010,Yao:2010:context-of-object-and-human,Bangpeng:2010:grouplet,sclaroff-eccv2010,VisualPhrases,Ferrari:2012:weakly_supervised_learning_interactions,delaitre-eccv2012}.  The idea is to use the person's appearance (body pose, hand shape, etc.) and the surrounding objects as mutual context---knowing the action helps predict the object, while knowing the object helps predict the action or pose.  For example, the Bayesian model in~\citep{gupta-pami2009} integrates object and action recognition to resolve cases where appearance alone is insufficient, e.g., to distinguish a spray bottle from a water bottle based on the way the human uses it.  Or, while it may be hard to infer body pose for a tennis serve, and hard to detect a tennis ball, attempting to do both jointly reduces ambiguity~\citep{Bangpeng:2010:grouplet}.  Similarly, structured models are developed to recognize manipulation actions~\citep{Kjellstrom:2008:simultaneous-recog-action-and-object} or sports activities~\citep{Yao:2010:context-of-object-and-human,desai-2010} in the context of objects.  Novel representations to capture subtle interactions, like playing vs.~holding a musical instrument, have also been developed~\citep{Bangpeng:2010:grouplet}.  Object recognition itself can benefit from a rich model of how human activity~\citep{peursum-iccv2005} or pose~\citep{delaitre-eccv2012} relates to the object categories.  While most such methods require object outlines and/or pose annotations, some work lightens the labeling effort via weakly supervised learning~\citep{sclaroff-eccv2010,Ferrari:2012:weakly_supervised_learning_interactions}.

While we are also interested in human-object interactions, our work differs from all the above in three significant ways.  First, whereas they aim to improve object or action recognition, our goal is to predict the location and size of an interactee---which, as we will show, has applications beyond recognition.  Second, we widen the definition of an ``interactee" to include not just manipulated objects, but also those that are untouched yet central to the interaction.  Third, and most importantly, the prior work learns the spatial relationships between the human and object in an \emph{action-specific} way, and is therefore inapplicable to reasoning about interactions for any action/object unseen during training.  In contrast, our approach is \emph{action-} and \emph{object-independent}; the cues it learns cross activity boundaries, such that we can predict where a likely interactee will appear even if we have not seen the particular activity (or object) before.
This means, for example, our method can learn that a person by a kitchen counter with arms outstretched is manipulating something around a certain size---whether he is making toast or blending juice.

\paragraph{Carried object detection}

Methods to detect carried objects (e.g.,~\citep{haritaoglu-pami2000,damen-eccv2008}) may be considered an interesting special case of our goal.  Like us, the intent is to localize an interactee object that (in principle) could be from any category, though in reality the objects have limited scale and position variety since they must be physically carried by the person.  However, unlike our problem setting, carried object detection typically assumes a static video camera, which permits good background subtraction and use of human silhouette shapes to find outliers.  Furthermore, it is specialized for a single action (carrying), whereas we learn models that cross multiple action category boundaries.

\paragraph{Social interactions}

Methods for analyzing social interactions estimate who is interacting with whom~\citep{cristani-bmvc2011,jimenez-gaze,fathi-social-interactions}, detect joint attention in video~\citep{park-cvpr2015}, predict where people are looking~\citep{recasens-nips2015}, or categorize how people are touching~\citep{ramanan-proxemics}.  The ``interactee" in our setting may be another person, but it can also belong to another object category.  Furthermore, whereas the social interaction work can leverage rules from sociology~\citep{cristani-bmvc2011} or perform geometric intersection of mutual gaze lines~\citep{jimenez-gaze,fathi-social-interactions}, our task requires predicting a spatial relationship between a person and possibly inanimate object.  Accordingly, beyond gaze, we exploit a broader set of cues in terms of body posture and scene layout, and we take a learning approach rather than rely only on spatial reasoning.
Furthermore, unlike any of the above, we study how person-centric saliency affects a third party's impression of what is important in an image.

\paragraph{Object affordances}

Methods to predict object affordances consider an object~\citep{saxena-affordance,desai-affordance} or scene~\citep{gupta-cvpr2011-workspace} as input, and predict which actions are possible as output.  They are especially relevant for robot vision tasks, letting the system predict, for example, which surfaces are sittable or graspable.  Our problem is nearly the inverse: given a human pose (and other descriptors) as input, our method predicts the localization parameters of the object defining the interaction as output.  %

\paragraph{Saliency and importance}

Saliency detection, studied for many years, also aims to make class-independent predictions about potentially interesting regions in an image.  While many methods look at low-level image properties (e.g.,~\citep{itti,Xiaodi:2007:saliency}), a recent trend is to \emph{learn} metrics for ``object-like" regions based on cues like convexity, closed boundaries, and color/motion contrast~\citep{liu-salient,Ferrari:2010:objectness,cpmc,endres-eccv2010,keysegments,Zitnick:2014:edgeBoxes}.
In object detection, these ``object proposal" methods have gained traction as a way to avoid naive sliding window search.  Related to our work, such metrics are category-independent by design: rather than detect a certain object category, the goal is to detect instances of \emph{any} object category, even those not seen in training.  However, neither saliency nor object proposals have been studied for identifying interactees, and, in contrast to our approach, none of the existing methods exploit person-centric cues to learn what is salient.

More related to our work are methods that model \emph{importance}~\citep{spain-eccv2008,spain-ijcv2011,sungju-bmvc2010,berg-importance}.  They attempt to isolate those objects within a scene that a human would be most likely to notice and mention.  Using compositional cues like object size and position~\citep{spain-eccv2008,spain-ijcv2011} as well as semantic cues about object categories, attributes, and scenes~\citep{berg-importance}, one can learn a function that ranks objects by their importance, or their probability of being mentioned by a human.  Similarly, a multi-view embedding between ranked word-lists and visual features can help retrieve images sharing prominent objects~\citep{sungju-bmvc2010}.  Like these methods, we aim to prioritize objects worth mentioning in a scene.  Unlike these methods, we propose a novel basis for doing so---the importance signals offered by a human-object interaction.  In addition, unlike methods that exploit object category-specific cues~\citep{berg-importance,sungju-bmvc2010,spain-eccv2008,spain-ijcv2011}, we learn a category-independent metric to localize a probable important object, relative to a detected person.

Different from any of the above saliency and importance work, our method predicts \emph{regions likely to contain an object involved in an interaction}.  We compare our method to representative state-of-the-art objectness~\citep{Ferrari:2010:objectness} and importance~\citep{berg-importance} metrics in our experiments, showing the advantages of exploiting human interaction cues when deciding which regions are likely of interest.

\paragraph{Describing images}

As one of the four applications of interactees, we explore image description.  Recent work explores ways to produce a sentence describing an image~\citep{Farhadi:2010,babytalk,yao-parsing-2010,Berg:2011:im2text,donahue-cvpr2015,larry-cvpr2015,karpathy-cvpr2015} or video clip~\citep{Saenko:2013:youtube2text}.  Such methods often smooth the outputs of visual detectors, making them better agree with text statistics~\citep{babytalk,Saenko:2013:youtube2text,infer-the-why} or a semantic ontology~\citep{yao-parsing-2010}.  One general approach is to produce a sentence by retrieving manually captioned images that appear to match the content of the novel query~\citep{Farhadi:2010,Berg:2011:im2text,devlin}.  Another is to employ language models to generate novel sentences~\citep{babytalk,larry-cvpr2015,Yejin:2012:generate-descriptions}.  Recently, methods based on multi-modal embeddings and deep learning show promise~\citep{karpathy-cvpr2015,donahue-cvpr2015,larry-cvpr2015}.  While a few existing methods employ human activity detectors~\citep{Berg:2011:im2text,Saenko:2013:youtube2text,infer-the-why}, they do not represent human-object interactions, as we propose.

Other methods explore various criteria for \emph{selectively} composing textual descriptions for images.  In~\citep{sadovnik-cvpr2012}, the system composes a description that best discriminates one image from others in a set, thereby focusing on the ``unexpected".  In~\citep{infer-the-why}, a language model is used to help infer a person's motivation, i.e., the purpose of their actions.  In~\citep{ordonez-iccv2013}, a mapping is learned from specific object categories to natural sounding entry-level category names (e.g., dolphin vs.~grampus griseus).  This notion of being selective about what to state in a description relates to our goal of selecting relevant content based on the interaction, but none of the prior work composes a description with an explicit model of interactions.

For experiments demonstrating interactees' potential to aid image description, we make use of a simple retrieval-based image description approach modeled after~\citep{Berg:2011:im2text,devlin}.  We do not claim new contributions for sentence generation itself; rather, we use this area as a testbed for exploiting our contribution of interactee localization.

\makeatletter{}%
\section{Approach}\label{sec:algorithm}

In the following, we first precisely define what qualifies as an interactee (Sec.~\ref{sec:define}) and interaction and describe our data collection effort to obtain annotations for training and evaluation (Sec.~\ref{sec:data}). Then, we explain the two proposed learning and prediction procedures, namely, our interaction embedding-based non-parametric approach (Sec.~\ref{sec:nonparam}) and a parametric probabilistic approach (Sec.~\ref{sec:param}).  Finally, we present the four tasks that exploit our method's interactee predictions (Sec.~\ref{sec:apps}).

\subsection{Definition of Human-Interactee Interactions}\label{sec:define}

An \emph{interactee} refers to the thing a particular person in the image is interacting with.  An interactee could be an object, a composition of objects, or another person.  To characterize interactees, then, first we must define precisely what a human-interactee \emph{interaction} is. This is important both to scope the problem and to ensure maximal consistency in the human-provided annotations we collect.

Our definition considers two main issues: (1) the interactions are not tied to any particular set of activity categories, and (2) an interaction may or may not involve physical contact.  The former simply means that an image containing a human-object interaction of any sort qualifies as a true positive; it need not depict one of a predefined list of actions (in contrast to prior work~\citep{Bangpeng:2011:stanford40,Everingham:2010:pascal-voc,gupta-pami2009,desai-2010,Yao:2010:context-of-object-and-human,sclaroff-eccv2010,Ferrari:2012:weakly_supervised_learning_interactions}).  By the latter, we intend to capture interactions that go beyond basic object manipulation activities, while also being precise about what kind of contact does qualify as an interaction.  For example, if we were to define interactions strictly by cases where physical contact occurs between a person and object, then walking aimlessly in the street would be an interaction (interactee=road), while reading a whiteboard would not.  Thus, for some object/person to be an interactee, the person (``interactor") must be paying attention to it/him and perform the interaction with a purpose.

Specifically, we say that an image displays a human-interactee interaction if either of the following holds:
\begin{enumerate}
\item The person is watching a specific object or person and paying specific attention to it.  This includes cases where the gaze is purposeful and focused on some object/person within 5 meters.  It excludes cases where the person is aimlessly looking around.
\item The person is physically touching another object/person with a specific purpose.  This includes contact for an intended activity (such as holding a camera to take a picture), but excludes incidental contact with the scene objects (such as standing on the floor, or carrying a camera bag on the shoulder).
\end{enumerate}

An image can contain multiple human-interactee relationships.  We assume each person in an image has up to one interactee.  This does not appear to be a limiting assumption for static image analysis, in that a person's attention is by definition focused on one entity at the moment their photograph is taken.  At test time, our method predicts the likely interactee location for each individual detected person in turn.

\begin{figure}[t]
\centering
\begin{tabular}{cc}
\subfigure[Instructions for crowdworkers for localizing an interactee in the image.]{
\centering
\fbox{\includegraphics[width=0.48\textwidth]{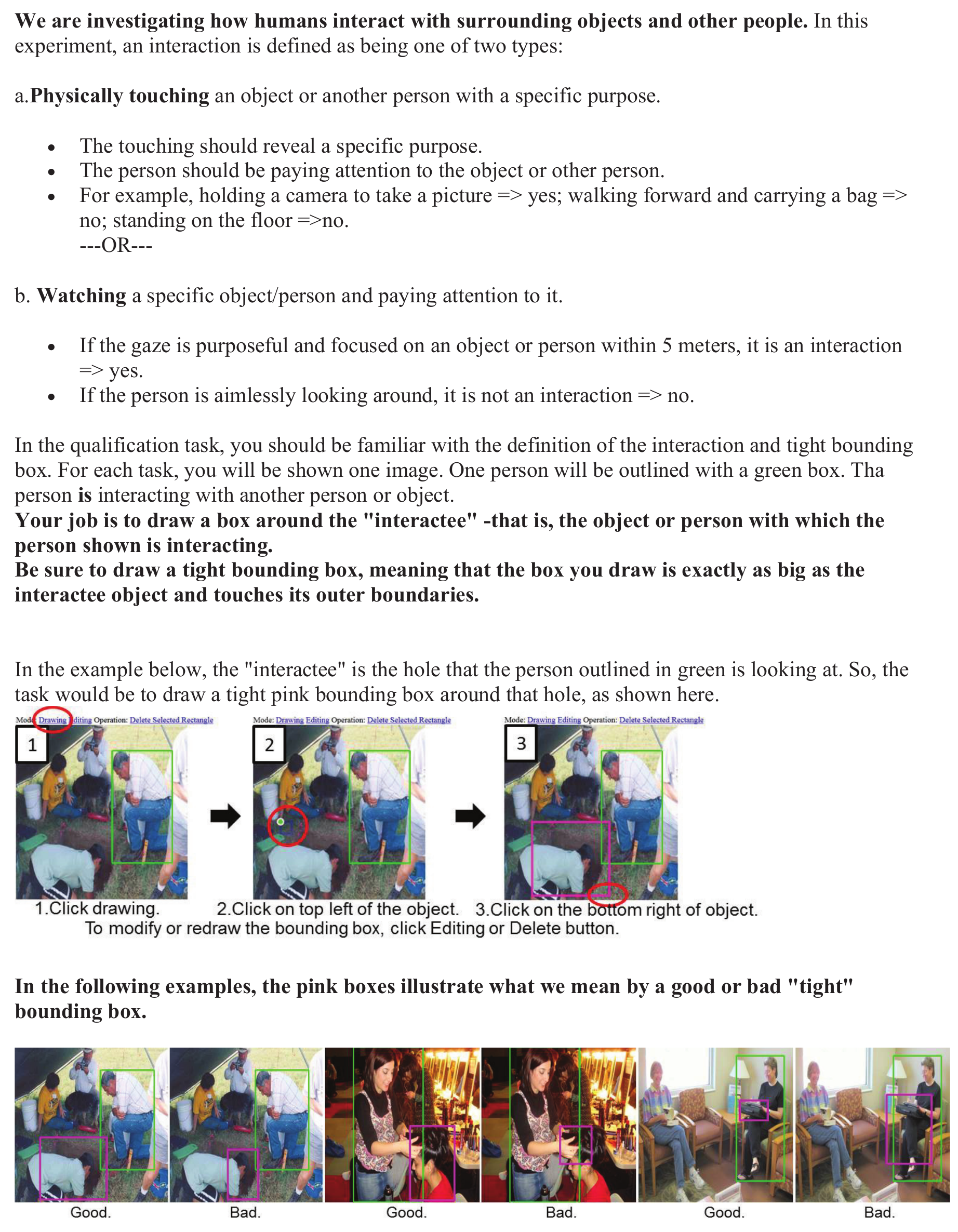}}}\\
\subfigure[Example task for localizing the interactee in an image.]{
\centering
\fbox{\includegraphics[width=0.48\textwidth]{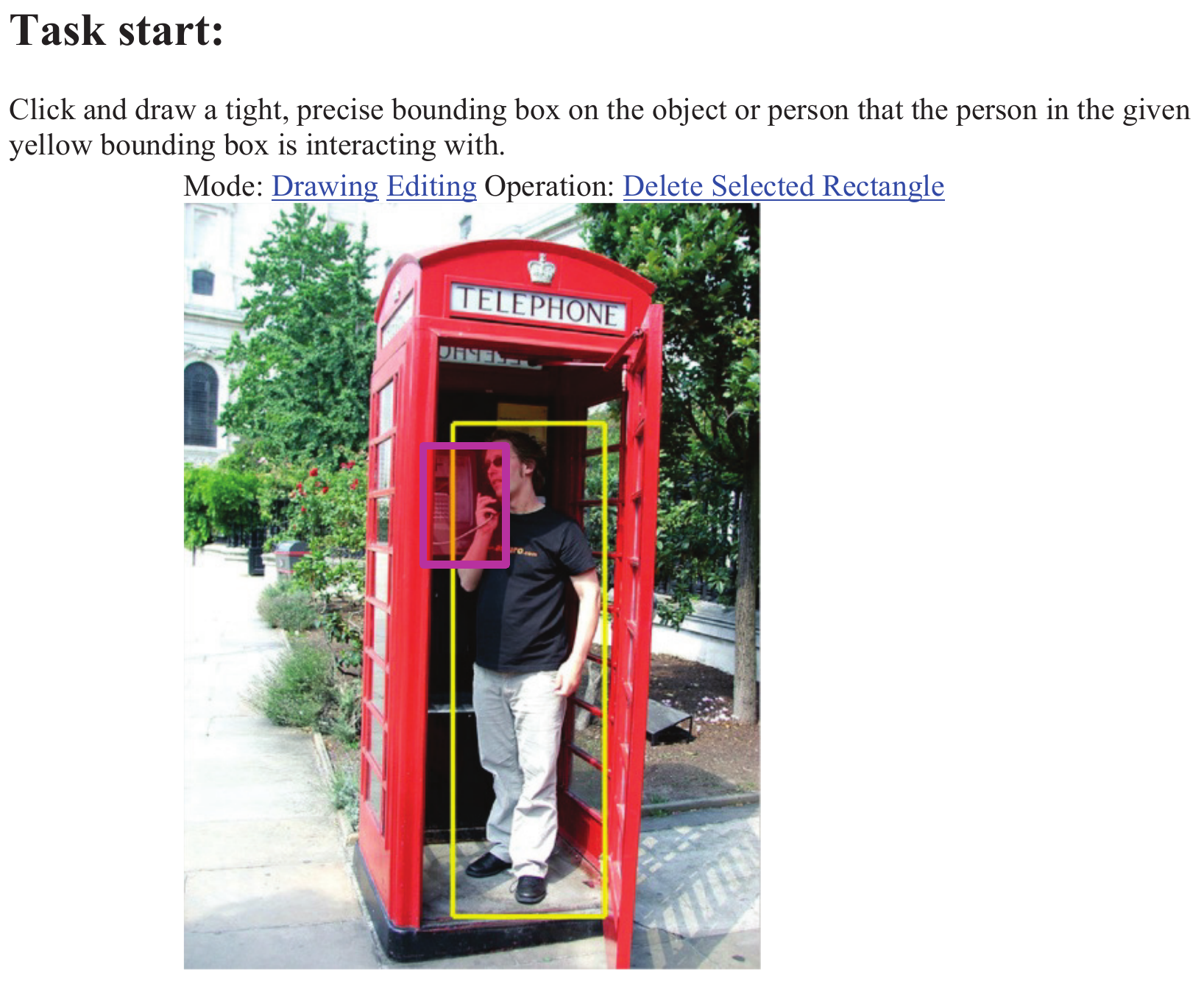}}}
\end{tabular}
\caption{Interface for interactee annotations via crowdsourcing.}
\label{fig:mturk-interactee-localization}
\end{figure}

\subsection{Interactee Dataset Collection}\label{sec:data}

Our method requires images of a variety of poses and interactee types for training.   We found existing datasets that contain human-object interactions, like the Stanford-40 and PASCAL Actions~\citep{Bangpeng:2011:stanford40,Everingham:2010:pascal-voc}, were somewhat limited to suit the category-independent goals of our approach.  Namely, these datasets focus on a small number of specific action categories, and within each action class the human and interactee often have a regular spatial relationship.  Some classes entail no interaction (e.g., \emph{running}, \emph{walking}, \emph{jumping}) while others have a  low variance in layout and pose (e.g., \emph{riding horse} consists of people in fairly uniform poses with the horse always just below).  While our approach would learn and benefit from such consistencies, doing so would essentially be overfitting, i.e., it would fall short of demonstrating action-independent interactee prediction.

Therefore, we curated our own dataset, the UT Interaction dataset, and gathered the necessary annotations.  We use selected images from three existing datasets, SUN~\citep{Xiao:2010:SUN}, PASCAL 2012~\citep{Everingham:2010:pascal-voc}, and Microsoft COCO~\citep{Lin:2014:coco}.  The selection is based solely on the need for the images to contain people.  SUN is a large-scale image dataset containing a wide variety of indoor and outdoor scenes.   Using all available person annotations\footnote{\texttt{http://groups.csail.mit.edu/vision/SUN/}}, we selected those images containing more than one person. The SUN images do not have action labels; we estimate these selected images contain 50-100 unique activities (e.g., \emph{talking}, \emph{holding}, \emph{cutting}, \emph{digging}, and \emph{staring}). PASCAL is an action recognition image dataset.  We took all images from those actions that exhibit the most variety in human pose and interactee localization---\emph{using computer} and \emph{reading}.  We pool these images together; our method does not use any action labels.  For COCO, we consider the subset of COCO training images that contain at least one person with an area exceeding 5,000 pixels and for which more than four out of five annotators report there is an interaction (see below). This yields a large number (more than 100) of unique activities including \emph{skiing}, \emph{skateboarding}, \emph{throwing}, \emph{batting}, \emph{holding}, etc.

\begin{figure*}[t]
\centering
\includegraphics[width=1.0\textwidth]{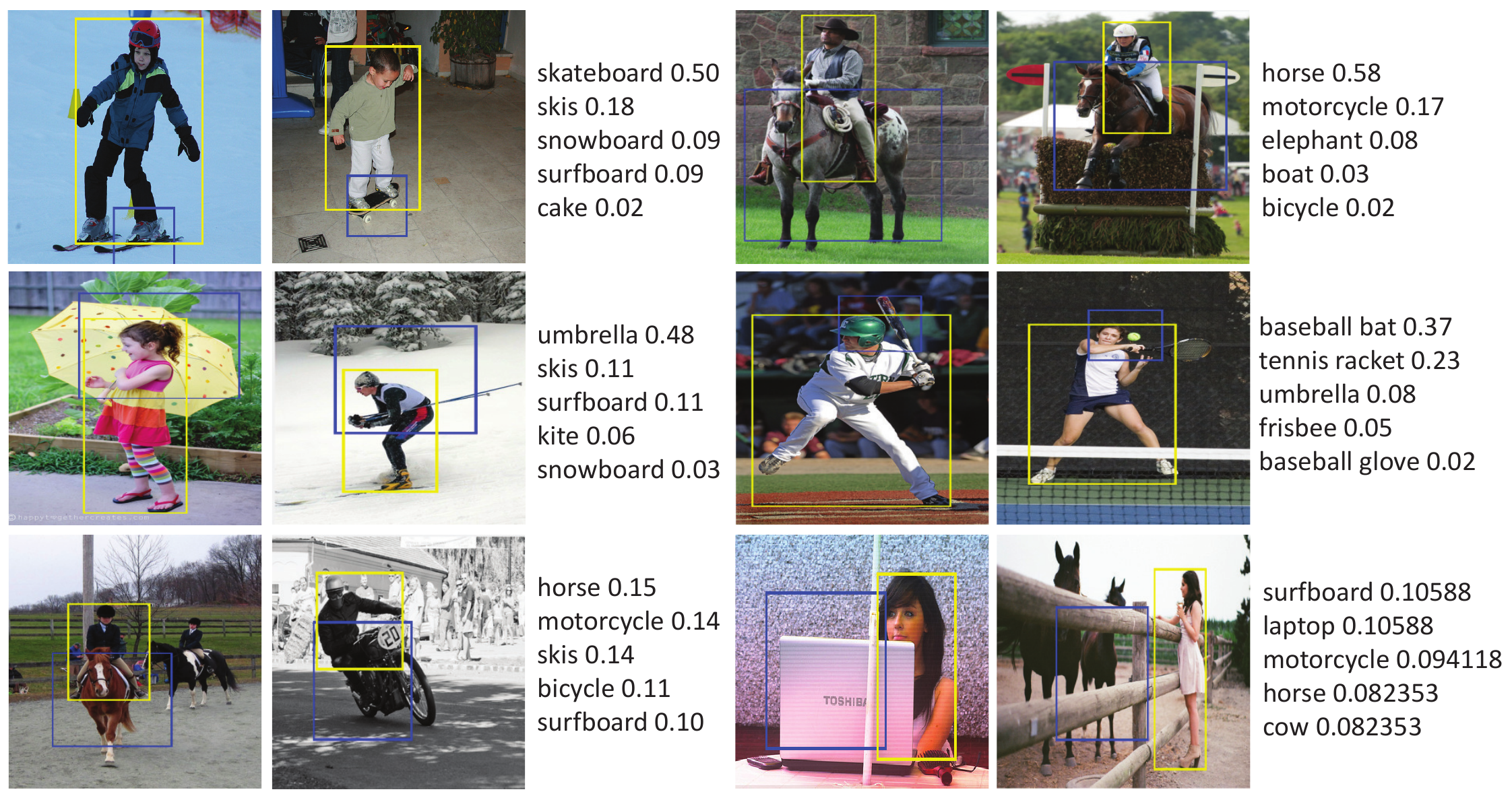}
   \caption[]{Examples for six of the 40 interaction types, as visualized with two image exemplars (left image pair in each example) plus the distribution over the interactee's category for all training images containing that discovered interaction type (right list of objects and probabilities for each example).  For legibility, we list only the top five categories with the highest probability for each distribution.  Yellow boxes indicate person region, blue boxes indicate ground truth interactee region.
     In examples in the top row, the distribution over object categories has lower entropy, meaning that the presence of that subject-interactee layout leaves relatively few possibilities for the category of the interactee.  As we move downward, the entropy of the distributions are higher.  The examples in the bottom row show cases where the object distribution has high entropy, meaning that the interaction type is often shared across different interactee categories.  See text for details.  } \label{fig:interaction-distribution}
\end{figure*}

We use Amazon Mechanical Turk (MTurk) to get bounding box annotations for the people and interactees in each image. Figure~\ref{fig:mturk-interactee-localization}(a) shows the instructions collecting the interactee localization in the form of bounding boxes.  We define what interaction means in our task, and we show examples of good localizations in the instructions. Next, we select images that are considered to have an interaction by most of  the annotators and collect the interactee's localization for them. Figure~\ref{fig:mturk-interactee-localization}(b) shows an example annotation task.

We get each image annotated by five to seven unique workers (due to the larger number of images in COCO, we have five unique workers for this dataset), and keep only those images for which at least  four workers said it contained an interaction.  This left 355/754/10,147 images from SUN/PASCAL/COCO respectively.

The precise location and scale of the various annotators' interactee bounding boxes will vary.  Thus, we obtain a single ground truth interactee bounding box via an automatic consensus procedure.  First, we apply mean shift to the coordinates of all annotators' bounding boxes.  Then, we take the largest cluster, and select the box within it that has the largest mean overlap with the rest.

The interactee annotation task is not as routine as others, such as tagging images by the objects they contain.  Here the annotators must give careful thought to which objects may qualify as an interactee, referring to the guidelines we provide them.  In some cases, there is inherent ambiguity, which may lead to some degree of subjectivity in an individual annotator's labeling.  Furthermore, there is some variability in the precision of the bounding boxes that MTurkers draw (their notion of ``tight" can vary).  This is why we enlist multiple unique workers on each training example, then apply the consensus algorithm to decide ground truth.  Overall, we observe quite good consistency among annotators.  The average standard deviation for the center position of bounding boxes in the consensus cluster is 8 pixels.  See Figure~\ref{fig:annotation-human-test-vs-gt}, columns 1 and 3, for examples.

\subsection{Localizing Interactees in Novel Images}\label{sec:localizing_interactee}

We explore two different methods for interactee localization: (1) a non-parametric regression approach that uses a learned interaction embedding and (2) a network-based probabilistic model.  %

In both methods, to capture the properties of the interactee in a category-independent manner, we represent its layout with respect to the interacting person.  In particular, an interactee's localization parameters consist of $\bm{y} = [x,y,a]$, where $(x,y)$ denotes the displacement from the person's center to the interactee box's center, and $a$ is the interactee's area.  Both the displacement and area are normalized by the scale of the person, so that near and far instances of a similar interaction are encoded similarly. Given a novel image with a detected person, we aim to predict $\bm{y}$, that person's interactee.

In the following, we first visualize the newly collected annotations in terms of the interactee types discovered (Sec.~\ref{sec:visualize}).  Then we overview the features our methods use for learning to predict interactees, including a deep learning based feature embedding (Sec.~\ref{sec:feats}).  Finally, we define the two models we investigate to make predictions based on those features (Secs.~\ref{sec:nonparam} and~\ref{sec:param}).

\subsubsection{Visualizing Discovered Interaction Types}\label{sec:visualize}

To visualize the types of interactions captured in our new annotations, we quantize the space of normalized interactee localization parameters $[x,y,a]$. Specifically, we apply k-means to $[x,y]$ and $[a]$ from training examples with $k=10$ and $k=4$ to get 10 and 4 clusters, respectively, for interactees' displacement and area. Next, we determine the interactee's location by its similarity to these clusters. This yields $T=10\times4=40$ \emph{interaction types}; images mapping to the same interaction type may contain different activities and objects, so long as the interactee is positioned and scaled in a similar manner relative to its ``subject" person.

Figure~\ref{fig:interaction-distribution} shows examples of six such interaction types, selected to convey the range of specificity present in the data.   For each type shown, we display two image examples and the distribution of interactee objects present across all training instances containing that interaction type.  Whereas the top row has relatively low entropy in that distribution, the bottom row has high entropy, and the middle row has entropy in between. This shows the spectrum of discovered interaction types in the data.  Some are fairly object-specific, where seeing the relative layout of subject and object already gives a strong prior about what object (interactee category) may be present.  For instance, in the top right example, the interaction layout suggests the person is riding something, whether a horse or a motorcycle.  Such cases naturally support the proposed application of interactees for object detector priming (defined below in Sec.~\ref{sec:apps-detect}), where we limit the application of an object detector to likely objects and likely places, based on the predicted interactee.  On the other hand, some interaction types are fairly object-independent, meaning that many object types are interchangeable given that person-object layout.  For instance, in the bottom right example, where an upright person is interacting with something of similar scale to his left, the interactee may be a surfboard, a laptop, a horse, etc.

\subsubsection{Interactee Descriptors}\label{sec:feats}

To learn the relationship between the interactee's location $\bm{y}$ and the image content, we extract three types of features.
We explore both learned and hand-crafted descriptors.

\begin{figure}[t]
\centering
\includegraphics[width=0.5\textwidth]{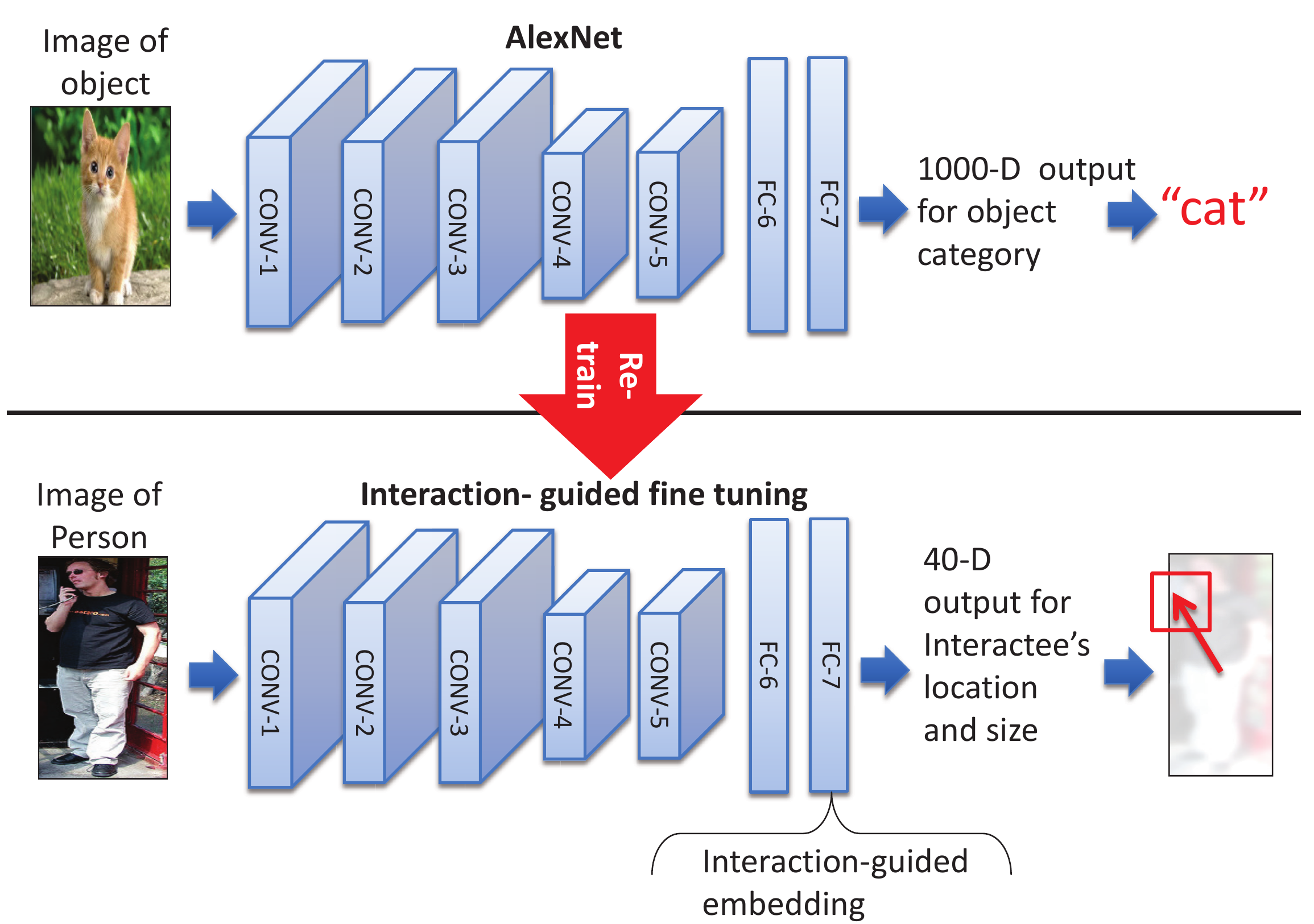}
   \caption[Interaction-guided fine-tuning]{Interaction-guided fine-tuning and network architecture of our interaction-guided embedding. We fine AlexNet~\citep{Alex:2012:imagenet}, which was originally trained for a 1000-way object recognition task (top), to revise the features to target our interactee localization task (bottom).  The 40-D output layer on the bottom corresponds to the $T=40$ interaction types discovered in the data via quantization of the interactee boxes' displacements and areas.}\label{fig:fine-tuning}
\end{figure}

\begin{figure}[t]
\centering
\fbox{\includegraphics[width=0.45\textwidth]{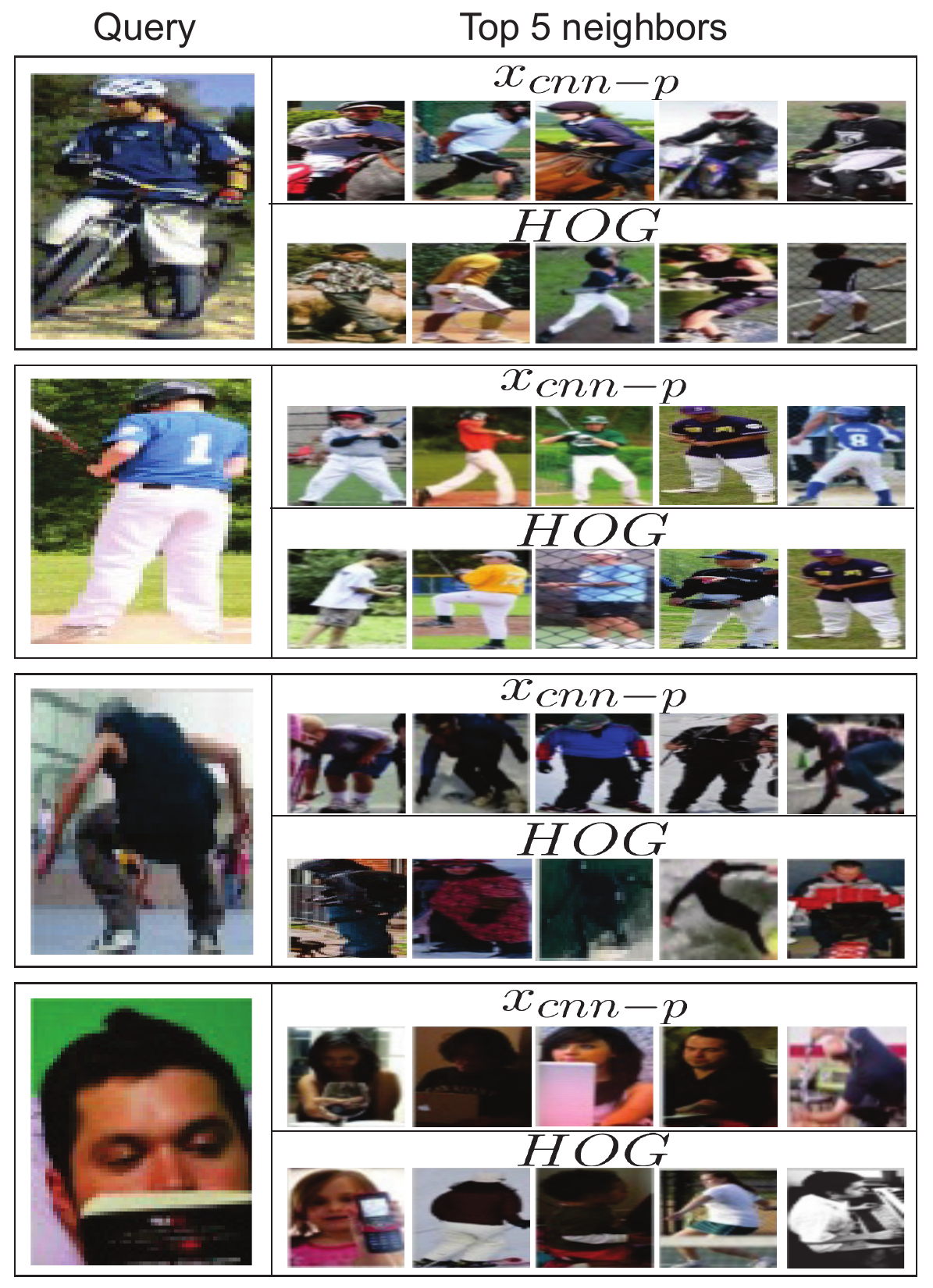}}
   \caption[Example of nearest pose neighbors by our $\bm{x}_{cnn-p}$ feature versus HOG feature]{Example of nearest pose neighbors by our $\bm{x}_{cnn-p}$ feature versus HOG feature. }\label{fig:pose-neighbors}
\end{figure}

First, we learn \emph{interaction-guided deep person features}.  Inspired by the idea that lower layer neurons in a CNN tend to capture the \emph{general} representation and the higher layer neurons tend to capture the representation \emph{specific} to the target task~\citep{Yosinski:2014:transferable_deep}, we fine-tune a deep convolutional neural network (CNN) for interactee localization.  In particular, as shown in Figure~\ref{fig:fine-tuning}, we use the quantization of the space of interactee localization parameters discussed above, then fine-tune a pre-trained object recognition network~\citep{Alex:2012:imagenet} to produce the proper discretized parameters when given a detected person (bounding box).  The last layer provides the learned feature map, $\bm{x}_{cnn-p}$.   This embedding discovers features informative for an interaction, which may include body pose cues indicating where an interactee is situated (e.g., whether the arms are outstretched, the legs close together, the torso upright or leaning, etc.), as well as attentional cues from the person's head orientation, eye gaze, or body position.

To illustrate qualitatively what the interaction-guided person descriptors capture, Figure~\ref{fig:pose-neighbors} shows some example nearest neighbor retrievals for a query image, whether using similarity in the learned feature space (top row of 5 images for each example) or using similarity in Histogram of Oriented Gradients (HOG)~\citep{hog} space.  Note that the more the retrieved examples correspond to the same localization of the interactee relative to the subject---even if \emph{not} the same object category---the better these results are.  The examples suggest that our $\bm{x}_{cnn-p}$ embedding is better able to represent precise interactions.

In a similar manner to the person embedding, we also learn \emph{interaction-guided deep scene features}.  We fine-tune a scene recognition network~\citep{Xiao:2014:place_cnn} to discover features about the entire scene that are useful for predicting the interaction, yielding an interaction-guided scene descriptor $\bm{x}_{cnn-s}$.  Intuitively, this embedding learns cues surrounding the person that are relevant to his interactee's placement, such as context for the activities that might be taking place.   It is also free to capture the appearance of the interactee itself (though due to the cross-category nature of interactions discussed above, this may or not be learned as useful.)

In the learned features defined thus far, the algorithm is free to learn anything about the scene and person that is informative for predicting the interactee localization correctly.  To further inject our domain knowledge about cues likely to be indicative of an interactee's placement, we next also consider an array of hand-crafted features.  In particular, we extract descriptors for \emph{pose}, \emph{scene layout}, and \emph{attention}, as follows.
\begin{itemize}
\item For the person's pose, we take the Histogram of Oriented Gradients (HOG)~\citep{hog} $\bm{x}_{h}$ computed in the person bounding box, plus the box's aspect ratio ($\bm{x}_a = \frac{h}{w}$) (e.g., the aspect ratio will be large for a standing person, smaller for a sitting person).
\item For scene layout, we take a GIST descriptor, $\bm{x}_g$, and the person's normalized position within the image, $\bm{x}_p$.  The latter captures how the person is situated within the scene, and thus where there is ``room" for an interactee.  For example, assuming a photographer intentionally framed the photo to capture the interaction, then if the person is to the far right, the interactee may tend to be to the left.
\item For attention, we use poselets~\citep{Maji:2011:poselet-action} to estimate the head and torso orientation, $\theta_{h}$ and $\theta_{t}$, to capture the direction of attention, whether physical or non-physical.  A poselet is an SVM that fires on image patches with a given pose fragment (e.g., a bent leg, a tilted head).  The head orientation offers coarse eye gaze cues, while the torso tells us which objects the person's body is facing.
\end{itemize}

Combining the learned and hand-crafted features, we have the complete feature vector
\begin{equation}
\bm{x} = [\theta_h, \theta_t, \bm{x}_h, \bm{x}_a, \bm{x}_g, \bm{x}_p, \bm{x}_{cnn-p}, \bm{x}_{cnn-s}].
\end{equation}

In Sec.~\ref{sec:experiment} we analyze the relative impact of the descriptors; the primary conclusion is that all contribute to interactee localization accuracy.

\subsubsection{Non-parametric Regression with Interaction-guided Embedding}\label{sec:nonparam}

Our first method for this task is simple and data-driven. It predicts the interactee in a novel image using the learned interaction-guided embedding together with non-parametric regression. %

We compute and store the descriptor above $\bm{x}$ for each interactee-annotated training image, yielding a set of $N$ training pairs
$\{(\bm{x}_i,\bm{y}_i)\}_{i=1}^N$.  To infer the interactee parameters
 \begin{equation}
 \bm{\hat{y}_q}=[\hat{x}_q,\hat{y}_q,\hat{a}_q]
  \end{equation}
for a novel query image $\bm{x}_q$, we use non-parametric locally weighted regression.  The idea is to retrieve training images most similar to $\bm{x}_q$, then combine their localization parameters.  Rather than simply average them, we attribute a weight to each neighbor that is a function of its similarity to the query.  In particular, we retrieve the $K$ nearest neighbors $\bm{x}_{n_1},\dots,\bm{x}_{n_K}$ from the training set based on their Euclidean distance to $\bm{x}_q$.  We normalize distances per feature by the standard deviation of the $L_2$ norms between training features of that type.  Then, the estimated interactee parameters are
\begin{equation}
\bm{\hat{y}_q} = \sum_{i=1}^K w_i \bm{y}_{n_i},
\end{equation}
where $w_i = \exp(- \| \bm{x}_q - \bm{x}_{n_i} \|)$.

Note that while interactees are a function of the action being performed, there is not a one-to-one correspondence.  That is, the same action can lead to different interactees (e.g., climb a \emph{tree} vs.~climb a \emph{ladder}), and vice versa (e.g., \emph{climb} a tree vs.~\emph{trim} a tree).  This supports our use of a category-independent spatial representation of the interactee.  Our method can benefit from any such sharing across verbs; we may retrieve neighbor images that contain people doing activities describable with distinct verbs, yet that are still relevant for interactee estimation.  For example, a person cutting paper or writing on paper may exhibit both similar poses and interactee locations, regardless of the distinct action meanings.  Thus, there is value here in not collapsing the dataset to verb-specific models.

A natural question is whether one could simply learn the localization parameters ``end-to-end" from the image rather than using the person/scene embeddings as an intermediary to a non-parametric learning approach.  In practice, we found such an approach inferior to ours.  This indicates there is value in 1) separating the person and scene during feature learning (more data would likely be needed if one wanted to treat the person as a latent variable) and 2) augmenting the learned features with semantically rich features like gaze.

\subsubsection{Probabilistic Model with Mixture Density Network}\label{sec:param}

We expect the non-parametric method described above to fare best when there is ample labeled data for learning. Since this is not always the case, as an alternative approach, we also consider a parametric model to represent interactee localization. In particular, we show how to  localize interactees using Mixture Density Networks (MDN)~\citep{Bishop:1994:mixturedensity} to build a predictive distribution for the interactee localization parameters. An MDN is a neural network that takes as input the observed features, their associated parameters, and as output produces a network able to predict the appropriate Gaussian mixture model (GMM) parameters for a novel set of observed features.

To build a predictive distribution for the interactee localization parameters, we want to represent a conditional probability density $P(\bm{y} | \bm{x})$. Here we model density as a mixture of Gaussians with $m$ modes:

\begin{equation}
P(\bm{y} | \bm{x}) = \sum_{i=1}^m\alpha_i\mathcal{N}(\bm{x}; \bm{\mu}_i, \bm{\Sigma}_i),
\end{equation} where $\alpha_i$ denotes the prior mixing proportion for component $i$, $\bm{\mu}$ is its mean, and $\bm{\Sigma}_i$ is its covariance matrix. We use the $N$ labeled training examples $\{(\bm{x}_i,\bm{y}_i)\}_{i=1}^N$ to train the MDN.

In testing, given a novel test image, we extract the descriptors from the person bounding box in the image.  Then, we use the learned MDN to generate the GMM $P(\bm{y}^t | \bm{x}^t)$  representing the most likely positions and scales for the target interactee.

In this way, we can assign a probability to any candidate position and scale in the novel image. To estimate the single most likely parameters, we use the center of the mixture component with the highest prior ($\alpha_i$), following~\citep{Bishop:1994:mixturedensity}.  The output interactee box is positioned by adding the predicted $(\hat{x},\hat{y})$ vector to the person's center, and it has side lengths of $\sqrt{\hat{a}}$.

\begin{figure}[t]
\centering
\includegraphics[width=0.5\textwidth]{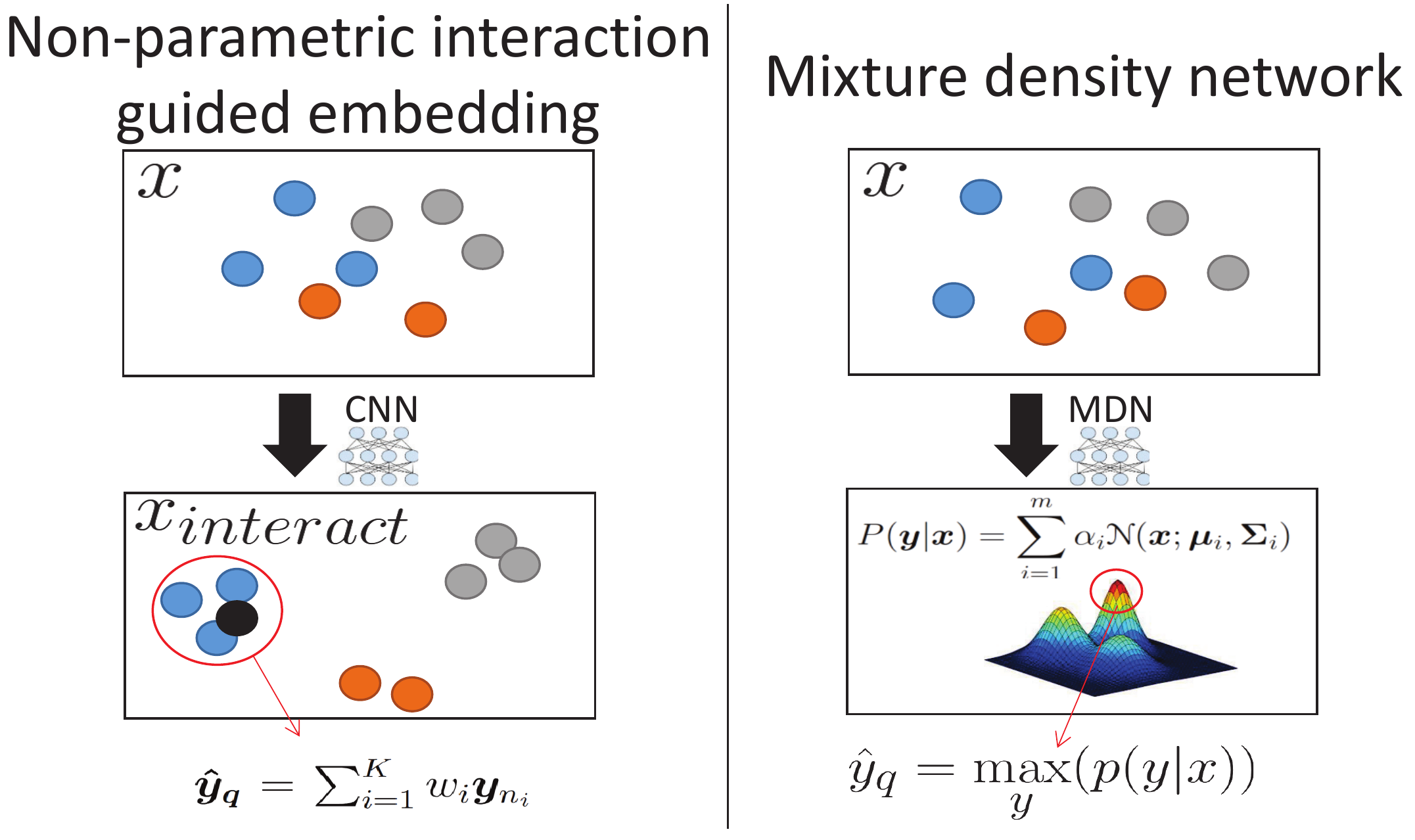}
   \caption{Comparison of the two learning approaches.  Similarly colored points denote image instances with similar interactee localization parameters.  See text for details.  }\label{fig:embedding-vs-cnn}
\end{figure}

Figure~\ref{fig:embedding-vs-cnn} recaps the two learning algorithms we consider for interactee localization.  The lefthand side depicts the non-parametric regression approach: given the training images in original image space (top), we learn CNN features that push those instances with similar interactee localizations together in the embedding space (bottom).  Then, a weighted combination of the labels for the nearest training instances provides the localization for a test image.  The righthand side depicts the MDN approach: given training images represented by the hand-crafted descriptors, we train a Mixture Density Network to produce Gaussian mixture model parameters for a distribution over likely localization parameters.  Then, the peak of that distribution provides the most likely localization for a test image.  These two pipelines primarily differ in terms of their (non)-parametric natures, as well as where learning is injected for the features.  For the former, learning is injected ``early", in that the feature space already captures the desired localization loss. In general we would expect the non-parametric approach to be more reliable, but more expensive at test time, when ample labeled data is available. On the other hand, the MDN approach has the advantage of allowing sampling of multiple sub-optimal solutions. We provide some analysis of their contrasts in Sec.~\ref{sec:experiment}.

\subsection{Applications of Interactee Prediction}\label{sec:apps}

Our method is essentially an object saliency metric that exploits cues from observed human-interactions.  Therefore, it has fairly general applicability.  To make its impact concrete, aside from analyzing how accurate its predictions are against human-provided ground truth, we also study four specific applications that can benefit from such a metric.

In the first task, we use the interactee localization to improve the accuracy or speed of an existing object detection framework by guiding the detector to focus on areas that are involved in the interaction (Sec.~\ref{sec:apps-detect}). In the second task, we use the interactee prediction to assist image retargeting. In this task, the image is resized by removing the unimportant content and preserving the parts related to the person and interactee (Sec.~\ref{sec:apps-seam}). In the third and fourth tasks, we explore how to leverage inferred interactees to detect important objects and generate image descriptions (Secs.~\ref{sec:importance} and~\ref{sec:sentence}).  These tasks aim to mimic human-generated image descriptions by focusing on the prominent object(s) involved in an interaction. Well-focused descriptions can benefit image retrieval applications, where it is useful to judge similarity not purely on how many objects two images share, but rather on how many \emph{important} objects they share.

\subsubsection{Task 1: Interactee-aware Contextual Priming for Object Detection}\label{sec:apps-detect}

First, we consider how interactee localization can prime an object detector.
As shown in Figure~\ref{fig:interaction-distribution}, a good interactee prediction can sometimes place a strong prior on which object detectors are relevant to apply, and where they are most likely to fire.  Thus, the idea is to use our method to predict the most likely place(s) for an interactee, then focus an off-the-shelf object detector to prioritize its search around that area.  This has potential to improve both object detection accuracy and speed, since one can avoid sliding windows and ignore places that are unlikely to have objects involved in the interaction.  It is a twist on the well-known GIST contextual priming~\citep{torralba-gist-priming}, where the scene appearance helps focus attention on likely object positions; here, instead, the cues we read from the person in the scene help focus attention.  Importantly, in this task, our method will look at the person, but will \emph{not} be told which action is being performed; this distinguishes the task from the methods discussed in related work, which use mutual object-pose context to improve object detection for a particular action category.

To implement this idea, we run the Deformable Part Model (DPM)~\citep{Felzenszwalb:2010:DPM} object detector on the entire image, then we apply our method to discard the detections that are outside the 150\% enlarged predicted interactee box (i.e., scoring them as $-\infty$). To alternatively save run-time, one could apply DPM to only those windows near the interactee.

\subsubsection{Task 2: Interactee-aware Image Retargeting}\label{sec:apps-seam}

As a second application, we explore how interactee prediction may assist in image retargeting.  The goal is to adjust the aspect ratio or size of an image without distorting its perceived content.  This is a valuable application, for example, to allow dynamic resizing for web page images, or to translate a high-resolution image to a small form factor device like a cell phone.  Typically retargeting methods try to avoid destroying key gradients in the image, or aim to preserve the people or other foreground objects.  Our idea is to protect not only the people in the image from distortion, but also their predicted interactees.  The rationale is that both the person and the focus of their interaction are important to preserve the story conveyed by the image.

To this end, we consider a simple adaption of the Seam Carving algorithm~\citep{Avidan:2007:seamcarving}.  Using a dynamic programming approach, this method eliminates the optimal irregularly shaped ``seams" from the image that have the least ``energy".  The energy is defined in terms of the strength of the gradient, with possible add-ons like the presence of people (see~\citep{Avidan:2007:seamcarving} for details).  To also preserve interactees, we augment the objective to increase the energy of those pixels lying within our method's predicted interactee box.  Specifically, we scale the gradient energy $g$ within both person and interactee boxes by $(g+5)*5$.

\subsubsection{Task 3: Interactees as Important Objects}\label{sec:importance}

A good visual recognition system ought to be able to parse and name every object in the scene.  An even \emph{better} recognition system would decide which among all the things it can recognize are even worth mentioning.  Thanks to rapid progress in recognition algorithms over the last 10 years, researchers are gradually shifting their focus to this next level of analysis.  In particular, exciting new developments show ways to map an image (or its detected visual concepts) to a natural language sentence~\citep{Farhadi:2010,babytalk,yao-parsing-2010,Berg:2011:im2text,Saenko:2013:youtube2text,donahue-cvpr2015,larry-cvpr2015,karpathy-cvpr2015}, or even explicitly rank the detected concepts by their perceived importance~\citep{spain-eccv2008,spain-ijcv2011,sungju-bmvc2010,berg-importance,sadovnik-cvpr2012}.  Methods that can concisely describe only the important parts of a scene will facilitate a number of interesting applications, including auto-captioning systems to assist the visually impaired, or image and video retrieval systems that index only the most important visual content.

\begin{figure}[t]
\centering
\includegraphics[width=0.5\textwidth]{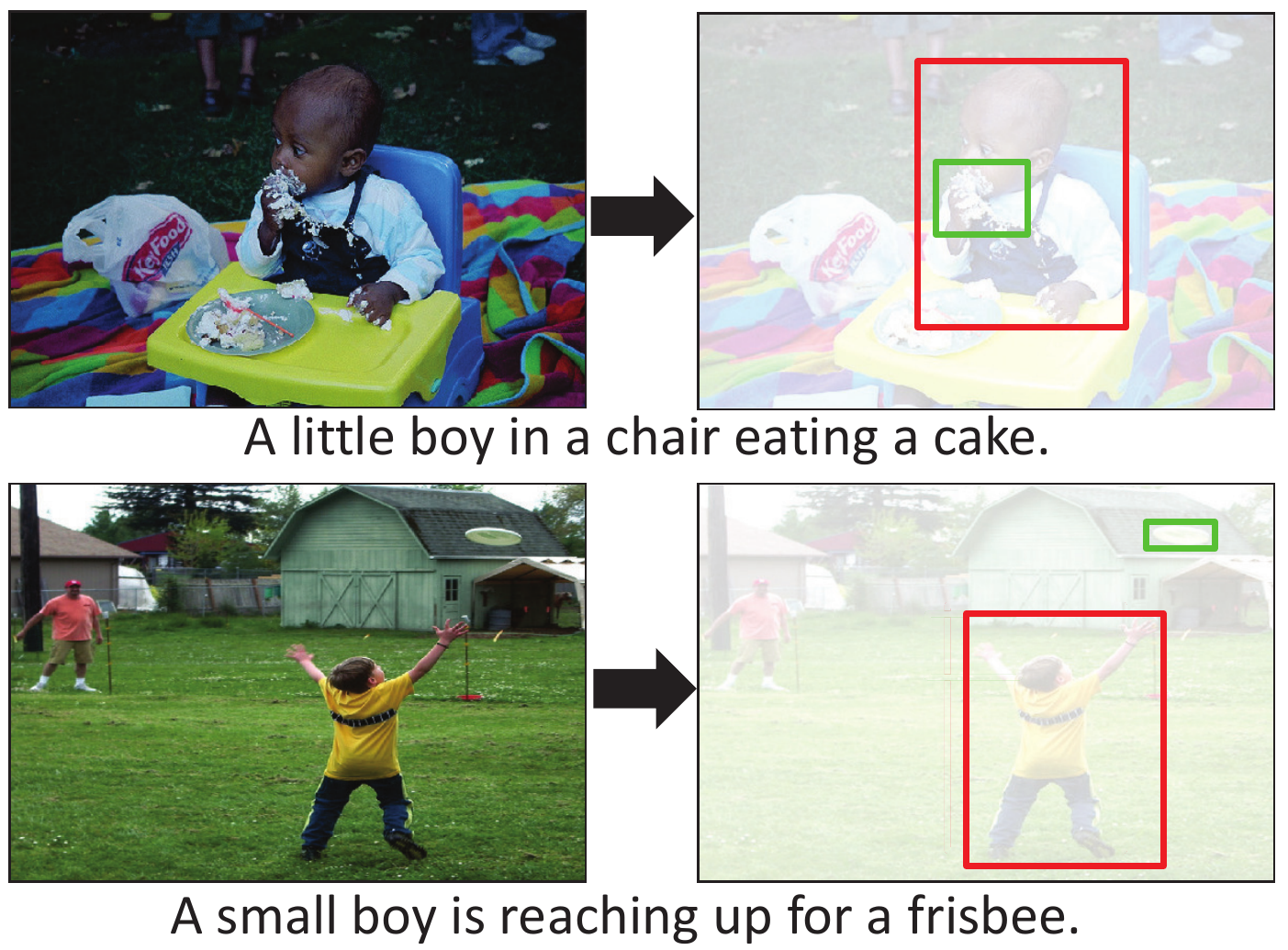}
   \caption[Describing an image with interactee]{When describing an image, people usually mention the object with which the person is interacting, even if it may be small or appear non-salient to traditional metrics.  For example, here the interactee objects are the cake and the frisbee. }\label{fig:sentence-concept}
\end{figure}

In the next and final two tasks, we apply interactees to explore the ``what to mention" question from a person-centric perspective. Given a novel image containing one or more people, how well does the interactee prediction indicate the objects in the scene are essential to generating an informative description?  As above, the key hypothesis is that a person's \emph{interactions} give vital cues.  For example, in Figure~\ref{fig:sentence-concept}, each image contains a dozen or more recognizable objects, but a human viewer has bias towards noticing the object with which each person interacts: the baby is eating \emph{cake} or the boy is reaching for the \emph{frisbee}.  Notably, not only do we focus on people and their activity---what they are doing, we also focus on the direct object of that activity---what they are doing t with/to.

In particular, in the third application we use interactees to gauge object \emph{importance} within a scene. Following prior work~\citep{spain-eccv2008,spain-ijcv2011,berg-importance}, we define ``important" objects as those mentioned by a human describing an image.  Our intuition that people tend to mention interactees is supported by data; in COCO, 80\% of true interactees appear in the human descriptions. Despite the fact that a person is the most commonly depicted object category in captioned images~\citep{Berg:2011:im2text}, existing methods to estimate object importance employ only generic compositional (like size and position) and semantic  (like the type of object or attribute) cues~\citep{spain-eccv2008,spain-ijcv2011,sungju-bmvc2010,berg-importance}.  The novelty of our approach to importance is to inject human-object interaction cues into the predictions.

We use predicted interactees to generate important object hypotheses, as follows.  Given a detected person, we project the predicted interactee bounding box (square box with the predicted area) into the query image.  This is essentially a saliency map of where, given the scene context and body pose, we expect to see an object key to the person's interaction.  Then, we sort all recognized objects in the scene by their normalized overlap with the interactee regions.  The first object in this list is returned as an important object.

\subsubsection{Task 4: Interactees in Sentence Generation}\label{sec:sentence}

Finally, in the fourth task, we generate complete sentences for the query image that account for its interactee.  While the importance task above focuses solely on the question of whether an object is important enough to mention, the sentence task entails also describing the activity and scene.

We take a retrieval-based approach, inspired by~\citep{Berg:2011:im2text,devlin}.  Again we use a non-parametric model.  Intuitively, if the content of a query image closely resembles a database image, then people will describe them with similar sentences.

Given a novel query, we compute $\bm{x}$ and its estimated interactee spatial parameters $\bm{\hat{y}}$, and use them together to retrieve the $K_s$ nearest images in a database annotated with human-generated sentences.  In particular, we use Euclidean distance to sort the neighbors, normalizing distances for $\bm{x}$ and $\bm{\hat{y}}$. Then, we simply adopt the sentence(s) for the query that are associated with those nearest examples.

We stress that our contribution is not a new way to infer sentences.  Rather, it is a new way to infer importance, which can be valuable to description methods.  Current methods for sentence generation~\citep{Farhadi:2010,babytalk,yao-parsing-2010,Berg:2011:im2text,donahue-cvpr2015,larry-cvpr2015,karpathy-cvpr2015,devlin}
are primarily concerned with generating a factually correct sentence; the question of ``what to mention" is treated only implicitly via text statistics. While we show the impact of our idea for retrieval-based sentence generation, it has potential to benefit other description algorithms too.  Arguably, once relevant high-level entities from visual processing are available (both object orderings as inferred by our method, as well as activities, scenes, etc.), the sentence generation step becomes a natural language processing problem.

\makeatletter{}%
\section{Experimental Results}\label{sec:experiment}

Our experiments evaluate three primary things: (1) how accurately do we predict interactees, compared to several baselines? (Sec.~\ref{sec:prediction-accuracy}), (2) how well can humans perform this task? (Sec.~\ref{sec:human}), and (3) the four applications of interactee localization (Sec.~\ref{sec:apps-experiment-result}).

\subsection{Datasets and Implementation Details}

\begin{figure*}[t]
\centering
\includegraphics[width=0.9\textwidth]{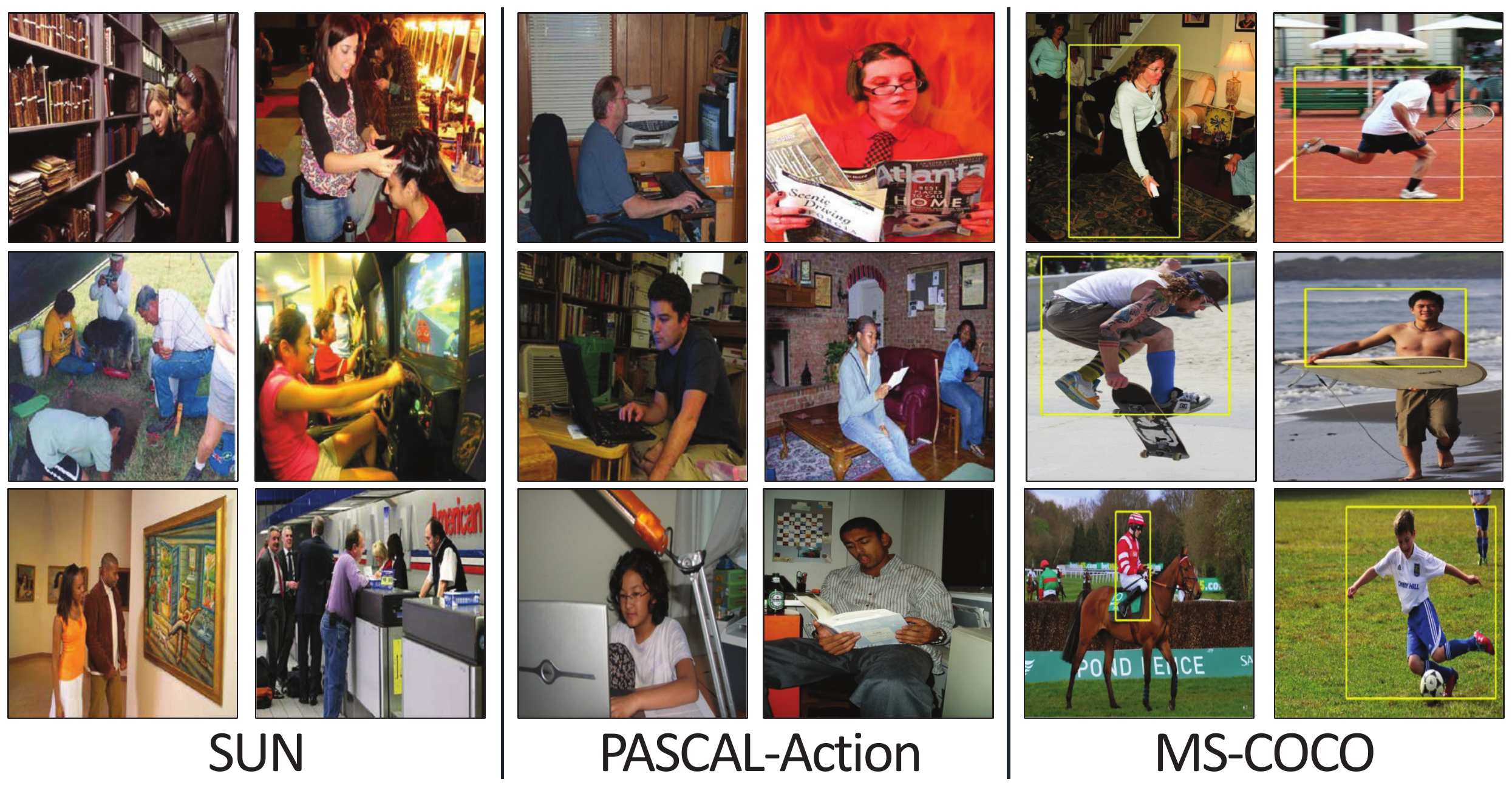}
   \caption[Examples of SUN, PASCAL-Action, and MS-COCO datasets]{Example images from the SUN, PASCAL-Action, and COCO datasets, which we get   annotated for interactions and interactees.} \label{fig:data-interactee}
\end{figure*}

We experiment with images containing people from three datasets: PASCAL Actions 2012~\citep{Everingham:2010:pascal-voc}, SUN~\citep{Xiao:2010:SUN}, and COCO~\citep{Lin:2014:coco}.  All three consist of natural, real-world snapshots with a wide variety of human activity. See Figure~\ref{fig:data-interactee} for example images originating from the three datasets.

For PASCAL and SUN, we use the subsets (cf.~Sec.~\ref{sec:data}) collated for human interactions, containing 754 and 355 images, respectively.  As PASCAL Actions and SUN do not have sentence data, we use them solely to evaluate interactee localization accuracy. For COCO, we use the 10,147 total images for which we obtained interactee bounding box annotations on MTurk (see Sec.~\ref{sec:data}).  COCO contains five human-written sentences per image, as well as object boundaries for 80 common object categories, which we exploit below.  For PASCAL and SUN we use a random 75\%-25\% train-test split and for COCO we use a random 80\%-20\% train-test split. We use the same train-test split for all our experiments.

For the feature embeddings, we fine-tune AlexNet~\citep{Alex:2012:imagenet} and Places-CNN~\citep{Xiao:2014:place_cnn} with the Caffe deep learning toolbox~\citep{Jia:2014:caffe}, for the deep person and scene features, respectively.  We use an SGD solver with 10,000 iterations and a learning rate of 0.001.  To form the target labels, we quantize the interactee's displacement and area into 10 and 4 bins, respectively, so the network provides $T=40$ outputs in the last layer.  We extract the features from the 7th layer (fc7) as $\bm{x}_{cnn-p}$ and $\bm{x}_{cnn-s}$ from each network. For HOG, each box is resized to $80\times80$ and we use cell size 8.

We localize interactee regions of interest automatically with our two proposed methods.  The inferred interactee localization guides us where to focus in the image for our four applications. We use annotated person boxes for our method and baselines to let us focus evaluation on the ``what to mention'' task, independent of the quality of the visual detectors.  We set $K=20$ and $K_s=5$ when retrieving the near neighbor interactions and images, respectively.  We fixed $K$ after initial validation showed values between 5-50 to perform similarly. For our embedding method, we use the hand-crafted features and fine-tuned interaction guided deep features. For our MDN method, we use the hand-crafted features only.\footnote{We tried to incorporate the deep feature for our MDN method, but the accuracy decreases due to the high feature dimensions that makes the GMM fitting harder.  Note that there are slight differences in Table~\ref{tab:allerrors} between the errors reported for our MDN method here and in the conference paper~\cite{chaoyeh-accv2014}.  This is because we have updated the hand-crafted features to be consistent with those used by the embedding method.}

\subsection{Accuracy of Interactee Localization}\label{sec:prediction-accuracy}

First we evaluate the accuracy of our interactee predictions.  Given an image, our system predicts the bounding box where it expects to find the object that is interacting with the person.  We quantify error in the size and position of the box. In particular, we report the difference in position/area between the predicted and ground truth boxes, normalized by the person's size. We also evaluate the accuracy of our method and baselines using the standard interaction over union (IOU) score between the inferred and ground truth interactee bounding boxes.

For core localization accuracy in this section, we compare to three baselines:
\begin{itemize}
\item The Objectness (which we abbreviate as Obj)~\citep{Ferrari:2010:objectness} method, which is a category-independent salient object detector.  Note that while our methods exploit cues about the person, the Objectness method is completely generic and does not.
\item A ``Near Person" baseline, which simply assumes the interactee is close to the person.  It predicts a box centered on the person, with a scale $\sim0.74$ of its area (a parameter set by validation on the training data).
\item A Random baseline, which randomly generates a position and size.
\end{itemize}
We score each method's most confident estimate in the below results.

\begin{table*}
\small
\centering
\resizebox{1.0\textwidth}{!}{
\begin{tabular}{|l|c||c|c|c|c|c|c|c|}
\hline
Metric   & Dataset &  Ours-embedding (w/CNN) & Ours-embedding (w/o CNN) & Ours-MDN & Obj~\citep{Ferrari:2010:objectness}           & Near Person   & Random\\
\hline \hline
\multirow{3}{*}{Position error}  & COCO    & \textbf{0.2256} & 0.2335 & 0.3058 & 0.3569 & 0.2909 & 0.5760 \\ \cline{2-8}
                                 & PASCAL  & \textbf{0.1632} & 0.1657 & 0.2108 & 0.2982 & 0.2034 & 0.5038 \\ \cline{2-8}
                                 & SUN     & 0.2524 & 0.2453 & \textbf{0.2356} & 0.4072 & 0.2456 & 0.6113 \\
\hline \hline
\multirow{3}{*}{Size error} & COCO        & \textbf{38.17} & 40.68 & 47.16 & 263.57 & 65.12 & 140.13 \\ \cline{2-8}
                            & PASCAL      & \textbf{27.04} & 28.95 & 36.31 & 206.59 & 31.97 & 100.31 \\ \cline{2-8}
                            & SUN         & \textbf{33.15} & 34.97 & 36.51 & 257.25 & 39.51 & 126.64 \\
\hline \hline
\multirow{3}{*}{IOU } & COCO        & \textbf{0.1989} & 0.1780 & 0.1153 & 0.0824 & 0.1213 & 0.0532 \\ \cline{2-8}
                      & PASCAL      & \textbf{0.2177} & 0.1998 & 0.1369 & 0.0968 & 0.1415 & 0.0552 \\ \cline{2-8}
                      & SUN         & \textbf{0.1710} & 0.1681 & 0.1145 & 0.1006 & 0.1504 & 0.0523 \\
\hline
\end{tabular}
}
\caption{Average interactee prediction performance as measured by position/size error and average IOU accuracy between prediction and ground truth interactee on all three datasets.}
\label{tab:allerrors}
\end{table*}

\begin{figure*}[t]
\centering
\includegraphics[width=1.0\textwidth]{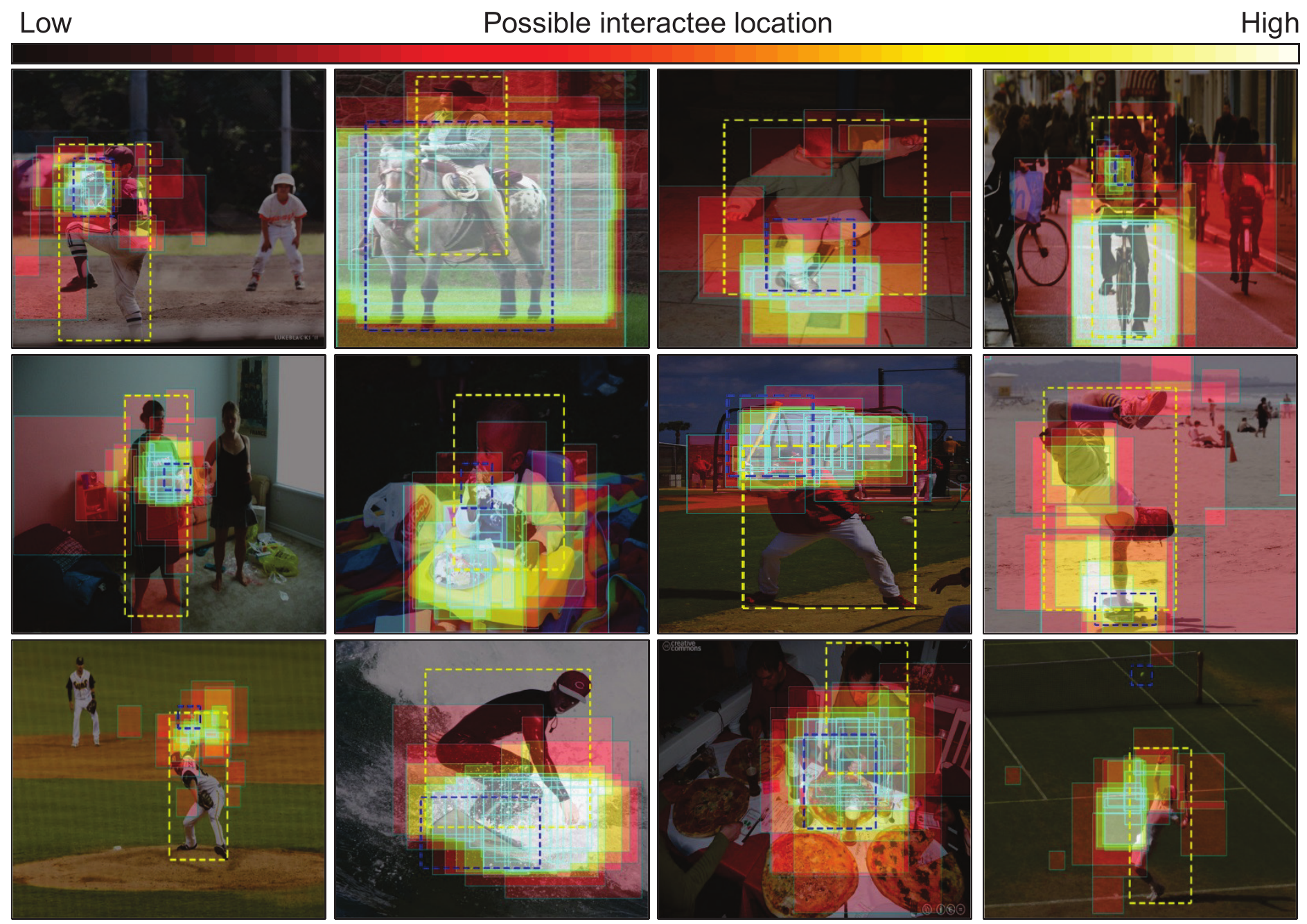}
   \caption[Example interactee localizations]{Example interactee localizations.  We display a heatmap for our embedding method's predictions by overlaying the retrieved training examples, such that they vote on likely areas of interest (white = high confidence).  The yellow dotted boxes indicate the main person in the image.  The blue box indicates the ground truth interactee location.  Our method can infer interactees in spite of varying interactions and object types.  The fourth column shows failure cases where there is less confidence in the prediction (see the upside down skater) or errors in unusual cases with multiple interactees (see the guy using the cell phone while riding a bike).  Best viewed in color.}\label{fig:heatmap}
\end{figure*}

Table~\ref{tab:allerrors} shows the result.  On three datasets, both of our methods offer significant improvement in position and size error over the baselines.  The margins are largest on the most diverse COCO dataset, where our data-driven approach (Ours-embedding) benefits from the large training set (COCO has more than 10 times the labeled instances than PASCAL or SUN). Our interaction embedding method provides 16\% lower errors over our MDN method on average. This indicates the strength of our learned features and data-driven estimation approach.  Our error reductions relative to Near Person average 17\% overall, and up to 22\% on COCO for object position.  However, on the SUN dataset, our MDN method is slightly better than our embedding method for interactee position; with only 355 images in SUN, our data-driven approach may suffer. Our gain over Near Person confirms that this is a non-trivial prediction task, particularly when the person is not touching the interactee (see the bottom example in third column in Figure~\ref{fig:heatmap}). As for the IOU metric, our embedding method provides significantly higher accuracy than other methods especially in COCO and PASCAL datasets with the help of larger data size. Our MDN method provides lower average IOU than the Near Person baseline due the low score cases from incorrect interactee localizations. For our non-parametric method, we also compare to variants which lack interaction-guided embedding features (Ours w/o CNN).  As shown in the Table, our fine-tuned features consistently improve the accuracy of our non-parametric method.

Figure~\ref{fig:heatmap} shows example predictions by the embedding variant of our method. We see that our method can often zero in on regions where the interaction is likely to be focused, even when the object may not have been seen in the training examples.  On the other hand, we also find failure cases, e.g., when a person's pose is too rare (upside down in the middle of fourth column) or the unusual cases with multiple interactees (using cell phone while riding bike in the top of fourth column).

\subsection{Human Subject Experiment}\label{sec:human}

\begin{figure}[t]
\centering
\includegraphics[width=0.5\textwidth]{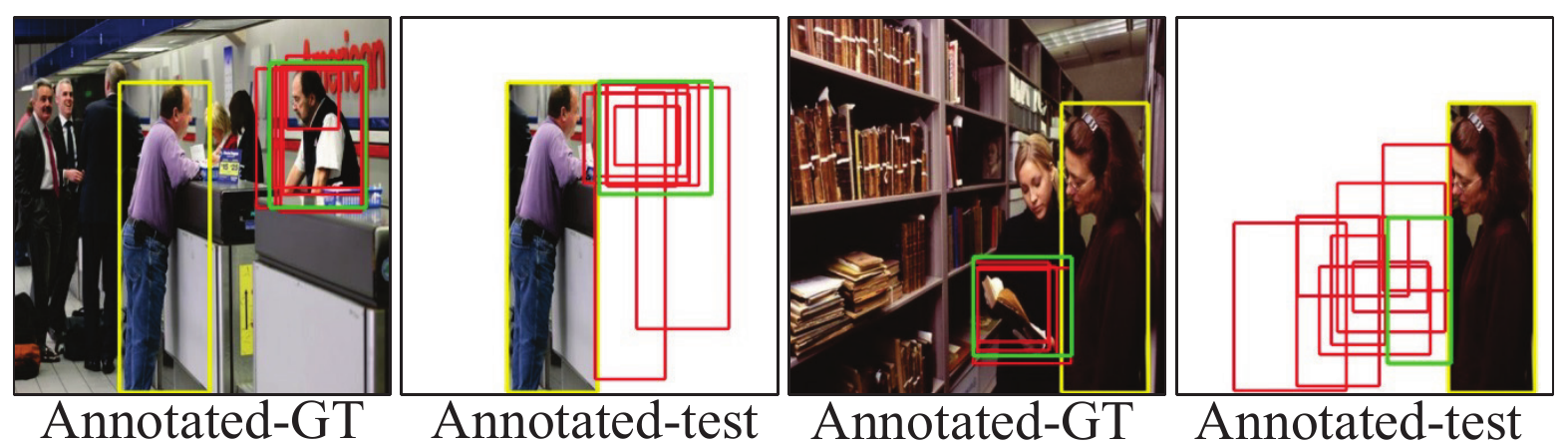}
   \caption[Example of human subject experiment]{We remove the background from the original image and ask human subjects to infer where the interactee might be. Red boxes denote their predictions, green box denotes consensus. Annotated-GT shows the full image (which is the format seen for ground truth collection, cf.~Sec.~\ref{sec:data}).  Annotated-test shows the human subject results. Naturally, annotators can more reliably localize the interactee when it is visible.}\label{fig:annotation-human-test-vs-gt}
\end{figure}

Next we establish an ``upper bound" on accuracy by asking human subjects on MTurk to solve the same task as our system. Note that when we collect the ground truth interactee localization, the annotators see the content of entire image. In this task, we remove the background from the original image and ask the human subjects to infer where the interactee might be. To have a fair comparison for the system, we use our MDN method with features extracted from only the person's bounding box: poselet feature~\citep{Maji:2011:poselet-action}, head/torso orientation, and the person's normalized position within the image, without knowing what category the interactee belongs to.  Then, we construct an interface forcing humans to predict an interactee's location with a similar lack of information.  Figure~\ref{fig:annotation-human-test-vs-gt}, columns 2 and 4, illustrate what the human subjects see, as well as the responses we received from 10 people.

Table~\ref{table:result-human-test-invisible-examples} shows the human subjects' results alongside the system's, for the subset of images in either dataset where the interactee is not visible within the person bounding box (since those cases are trivial for the humans and require no inference). The humans' guess is the consensus box found by aggregating all 10 responses with mean shift as before.  The humans have a harder time on SUN than PASCAL, due to its higher diversity of interaction types.  This study elucidates the difficulty of the task.  It also establishes an (approximate) upper bound for what may be achievable for this new prediction problem.  Note that the numbers in Table~\ref{table:result-human-test-invisible-examples} are not the same as those in Table~\ref{tab:allerrors} because 1) the test set is small here to focus on cases where the interactee does not overlap the person, and 2) we are depriving the system of the full scene features to make the test consistent with the human subject test.

\begin{table}[t]
\small
\centering
\resizebox{0.5\textwidth}{!}{
\begin{tabular}{|l|c|c|c|c|c|c|}
\hline
   &  \multicolumn{3}{c|}{Human subject}    & \multicolumn{3}{c|}{Ours-MDN}  \\
\hline
            & Position error & Size error & IOU & Position error & Size error & IOU\\
\hline
SUN w/o visible      & $0.1625$ & $29.61$ & $2.2768$ & $0.2767$ & $32.33$ & $0.0964$\\
\hline
PASCAL w/o visible   & $0.1035$ & $42.09$ & $0.3935$ & $0.2961$ & $43.27$ & $0.1320$ \\
\hline
\end{tabular}
}
\caption[Results of the human subject test]{Results of the human subject test.  This study demonstrates the difficulty of the learning task, and gives a rough upper bound for what an approach looking solely at the person in the image (but not the rest of the scene) could potentially achieve.}
\label{table:result-human-test-invisible-examples}
\end{table}

\subsection{Results for Applications of Interactee Prediction}\label{sec:apps-experiment-result}

Finally, we evaluate our idea in the context of the four tasks defined above.

\subsubsection{Task 1: Interactee-aware object detector contextual priming}\label{sec:improve-detection}

We first demonstrate the utility of our approach for contextual priming for an object detector, as discussed in Sec.~\ref{sec:apps-detect}, Task 1.  We use the PASCAL training images to train DPMs to find computers and reading materials, then apply our methods and the baselines to do priming.  We experiment with these two objects because the interactions between them and the person are more diverse than other types of objects in the dataset (for example, to detect a horse in riding-related images, the area below the person is always correct).

\begin{figure}[t]
\centering
\begin{tabular}{c}
\subfigure[Using computer]{
\includegraphics[width=0.24\textwidth]{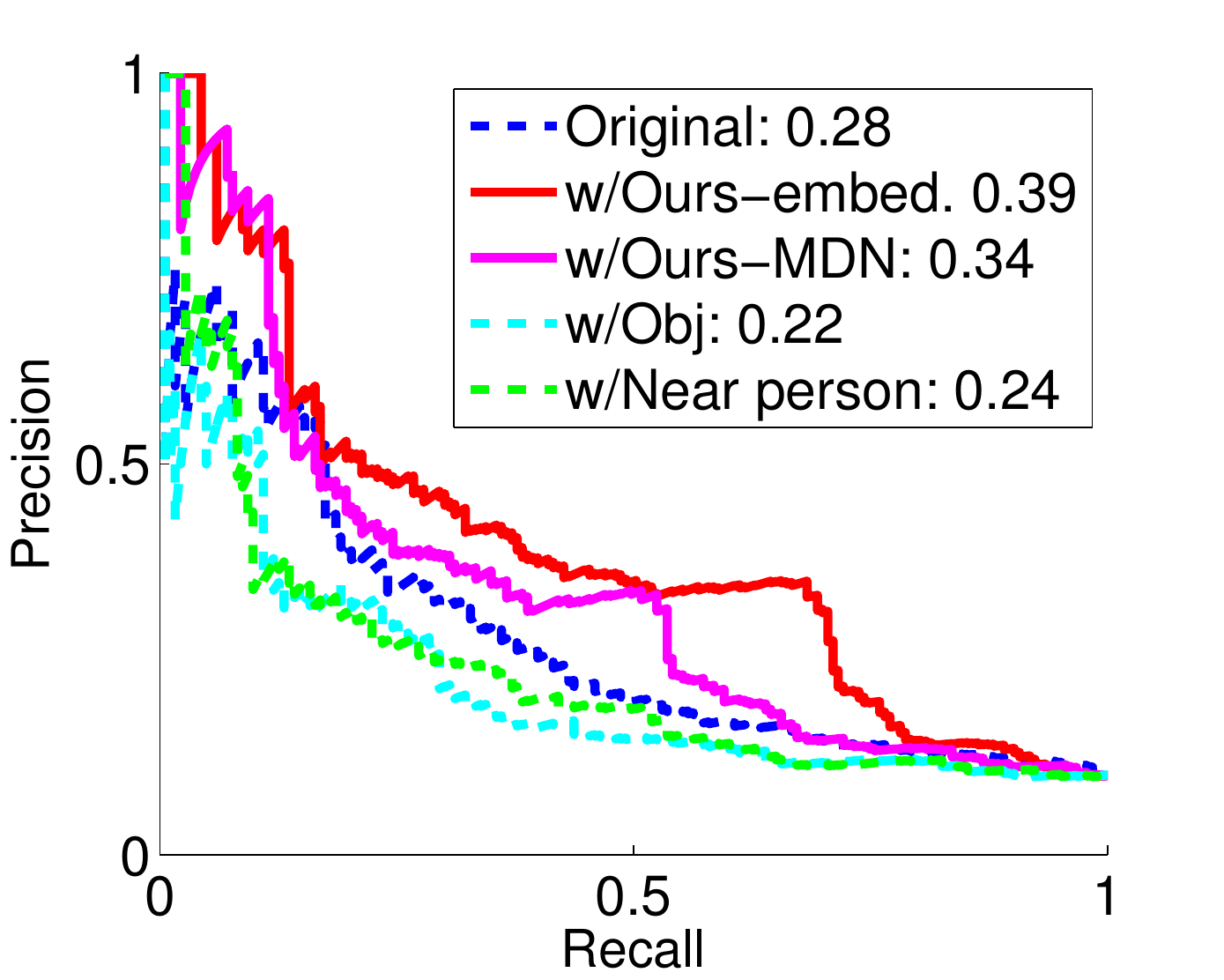}}
\subfigure[Reading]{
\includegraphics[width=0.24\textwidth]{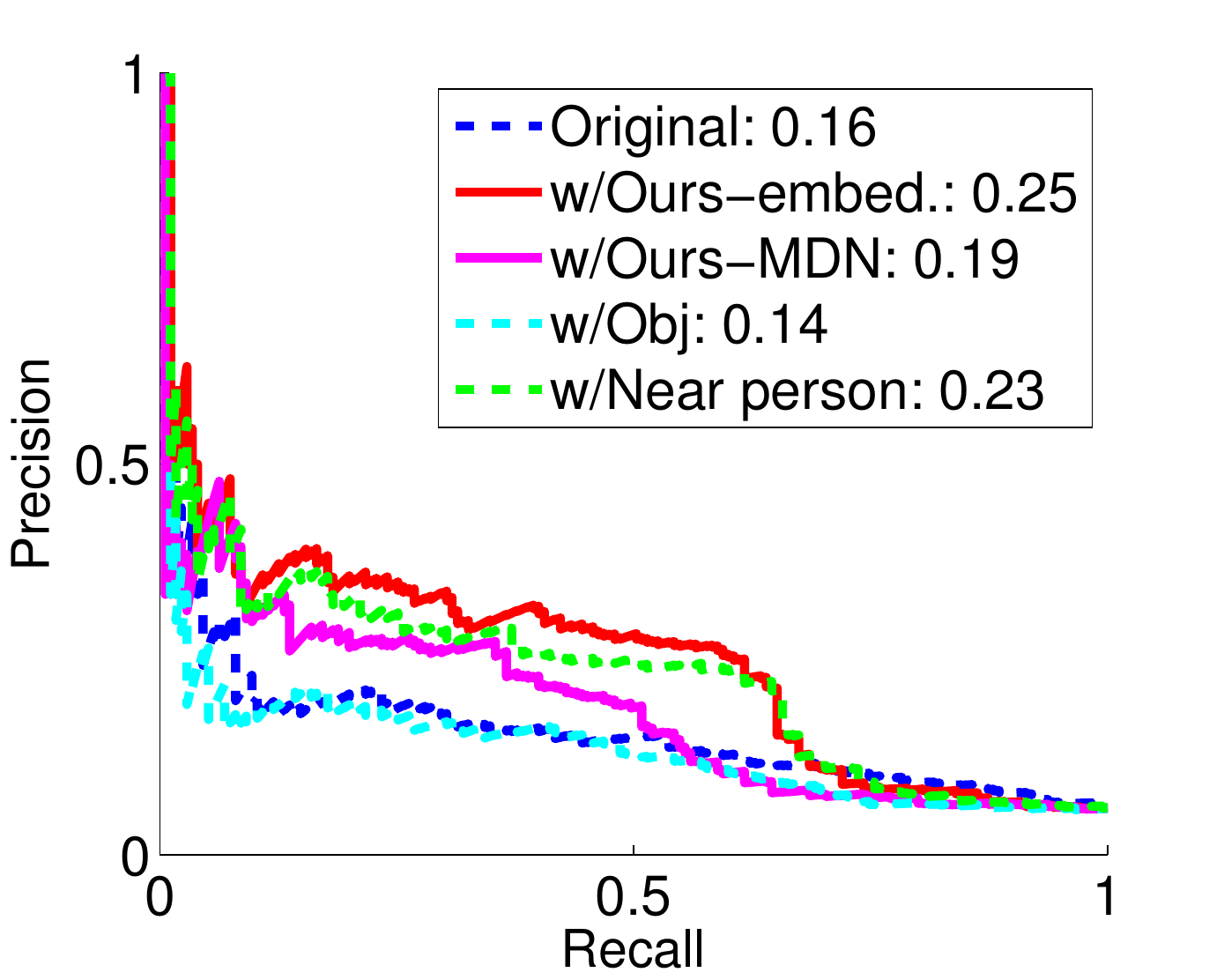}}
\end{tabular}
   \caption[Interactee context helps focus the object detector]{Using the interactee prediction as context helps focus the object detector.  Numbers in legends denote mean average precision (mAP).}
\label{fig:improve-dpm-cropping}
\end{figure}

Figure~\ref{fig:improve-dpm-cropping} shows the results.  We see our methods outperform the baselines, exploiting what they infer about the person's attention to better localize the objects. Note that neither of our methods use the action category labels during training. As was the case for pure localization accuracy above, our interaction embedding method again outperforms our MDN method.  The MDN approach remains substantially better than all the baselines on computer, though it underperforms the Near Person baseline on reading materials.  In that case Near Person fares well for the \emph{reading} instances because the book or paper is nearly always centered by the person's lap.

This experiment gives a proof of concept on challenging data that using the proposed interactee predictions can improve a standard object detection pipeline by knowing where it is most fruitful to look for an interacting object.

\begin{figure*}[t]
\centering
\includegraphics[width=1.0\textwidth]{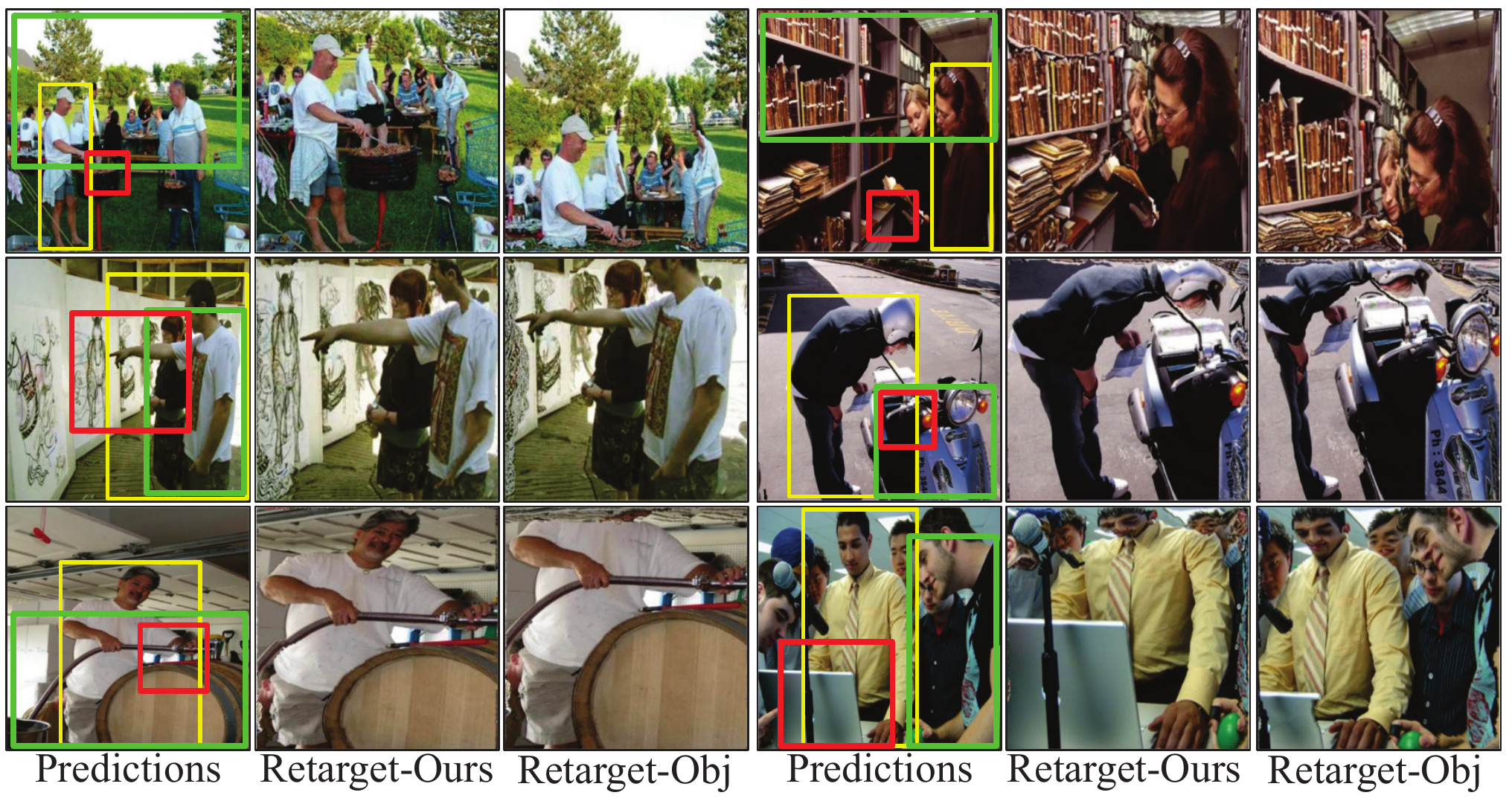}
   \caption[Interactee-aware image retargeting example results]{Interactee-aware image retargeting example results.  Our method successfully preserves the content of both the interactee (e.g., BBQ kit, book, painting of horse, laptop) and person, while reducing the content of the background. Objectness cannot distinguish salient objects that are and are not involved in the activity, and so may remove the informative interactees in favor of background objects.  The bottom right example is a failure case for our method, where our emphasis on the interactee laptop looks less pleasing than the baseline's focus on the people.}\label{fig:seam-carving-example}
\end{figure*}

\subsubsection{Task 2: Interactee-aware image retargeting}\label{sec:retargeting-by-seam-carving}

Next, we inject our interactee predictions into the Seam Carving retargeting algorithm, as discussed in Sec.~\ref{sec:apps-seam}, Task 2.  Figure~\ref{fig:seam-carving-example} shows example results.  For reference, we also show results where we adapt the energy function using Objectness's top object region prediction.  Both methods are instructed to preserve the provided person bounding box.  We retarget the source $500 \times 500$ images to $300 \times 300$.

We see that our method preserves the content related to both the person and his interactee, while removing some unrelated background objects.  In contrast, Objectness~\citep{Ferrari:2010:objectness}, unaware of which among the prominent-looking objects might qualify as an interactee, often discards the interactee and instead highlights content in the background less important to the image's main activity.

\subsubsection{Task 3: Interactees as important objects}\label{sec:mentioned_object_prediction}

Next, we use the interactee region of interest to predict object importance (cf.~Sec.~\ref{sec:importance}, Task 3). Following~\citep{spain-eccv2008,berg-importance}, we are given an image plus a list of objects and their categories/locations.  Ground truth importance is judged by how often humans mention the object in a caption.  Here we use the COCO data, since it comes with sentence annotations and ground truth object outlines.

For this task we compare to the existing \emph{Object Prediction} importance method of~\citep{berg-importance} (Sec.~4.1 in that paper).  It trains a logistic regression classifier with features based on object size, location, and category. To ensure fair comparison, we use the COCO data to train it to predict the object most often mentioned in the image. We again compare to the Near Person baseline, and two additional baselines:
\begin{itemize}
\item Prior, which looks at all objects present in the image and picks the one most frequently mentioned across all training images
 \item Majority, which predicts people will mention the object that happens most frequently in the test image.
\end{itemize}

All methods ignore the persons in the images, since all images have a person.  For this result, we discard images with only a person and a single object since all methods can only output that same object, leaving 1,617 test images.

Table~\ref{tab:mentioned_count} shows the result of 10 train/test splits.  We measure accuracy by the hit rate---the average percentage of ground truth sentences mentioning the object deemed most important, per image.  If each of the 5 ground truth captions include the predicted object, the score is 100\% for that image.  First, we see that interactees are correlated with important objects; the ground truth interactee leads to a hit rate of $78.4$.  Furthermore, our embedding method predictions outperform the baselines.  The nearest competing method is Near Person.  Even though the region of interest it predicts is substantially less precise (as we saw in Table~\ref{tab:allerrors}), it does reasonably well because the step of identifying the annotated COCO object nearest to that region is forgiving.  In other words, even a region of interest that is only partially correct in is localization can have the desired effect in this application, if it has no greater overlap with any of the other annotated COCO objects. Nonetheless, the ground truth upper bound reinforces that better precision does translate to better performance on solving this task. Similar to Sec~\ref{sec:prediction-accuracy}, our non-parametric method outperforms our MDN method.

The state-of-the-art importance method~\citep{berg-importance} is less accurate than our interactee-based method on this data.  We think this is because in the COCO data, an object of the same category, size, and location is sometimes mentioned, sometimes not, making the compositional and semantic cues used by that method insufficient.  In contrast, our method exploits interactions to learn if an object would be mentioned, independent of its position and category.  This result does not mean the properties used in~\citep{berg-importance} are not valuable; rather, in the case of analyzing images of people involved in interaction, they appear insufficient if taken alone.

\begin{table}[t]
\footnotesize
\centering
\begin{tabular}{|l|c|c|c|c|}
\hline
Method & Mention rate (\%) \\
\hline
Ground truth interactee & 78.4 (0.6) \\
\hline
\hline
Ours-embedding & \textbf{70.5 (0.4)} \\
\hline
Importance~\citep{berg-importance} & 65.4 (0.4) \\
\hline
Ours-MDN & 65.2 (0.5) \\
\hline
Near Person & 67.5 (0.5) \\
\hline
Prior & 64.6 (0.6) \\
\hline
Majority &  51.7 (0.6) \\
\hline
\end{tabular}
\caption[Average hit rates for predicted important objects]{Average hit rates (higher is better) for predicted important objects.  Numbers in parens are standard errors.}
\label{tab:mentioned_count}
\end{table}

\begin{table*}[t]
\scriptsize
\centering
\resizebox{1.0\textwidth}{!}{
\begin{tabular}{|l|c|c|c|c||c|}
\hline
                 &   1-Gram BLEU & 2-Gram  BLEU  & 3-Gram BLEU   & 4-Gram BLEU & Combined BLEU\\
\hline
Random                  &   55.19   & 19.26     & 4.18    & 1.26    & 8.65 \\
\hline
Global Matching~\citep{Berg:2011:im2text}      &   63.80   & 28.02     & 9.80   & 3.75 & 16.01    \\
\hline
Global+Content Matching~\citep{Berg:2011:im2text}(Actions$^\ast$) & 63.19   & 27.12     & 9.13    & 3.41  & 15.20 \\
\hline
Global Matching+AlexNet fc7 & 68.21 & 33.38 & 13.32 & 5.44 & 20.16 \\
\hline
Global Matching+Places-CNN  & 69.01 & 33.99 & 13.87 & 6.77 & 20.81 \\
\hline
\hline
Ours-embedding w/o cnn feature & 65.08 & 29.74 & 11.13 & 4.56 & 17.64 \\
\hline
Ours-embedding w/cnn-p only & 68.03 & 33.30 & 13.45 & 5.64 & 20.36 \\
\hline
Ours-embedding w/cnn-s only & 70.78 & 36.42 & 15.96 & 6.87 & 23.05 \\
\hline
Ours-embedding w/all & \textbf{73.85} & \textbf{40.33} & \textbf{18.88} & \textbf{8.68} & \textbf{26.43} \\
\hline
\end{tabular}
}
\caption[Average BLEU scores between query and retrieved sentences]{Average BLEU scores between query and retrieved sentences (higher = more similar).  See text for details. }
\label{tab:sentence-retrieval}
\end{table*}

\subsubsection{Task 4: Interactees in sentence generation}\label{sec:sentence-result}

Finally, we study how interactee detection can benefit retrieval-based sentence generation (cf.~Sec.~\ref{sec:sentence}, Task 4).
For each test image, we retrieve $K_s=5$ images from the training set, then compute the average similarity between the ground truth query and training sentences.  We use the standard BLEU score~\citep{Papineni:2002:BLEU} for $n$-gram overlap precision.

We compare our interaction embedding based regression approach to a retrieval-based sentence generation method in prior work~\citep{Berg:2011:im2text}.  For~\citep{Berg:2011:im2text}, there are two variations: Global Matching, which retrieves neighbors based on GIST and Tiny Image descriptors, and Global$+$Content Matching, which reranks that shortlist with the local image content as analyzed by visual detectors.  The methods lack publicly available code, so we implement them ourselves.  The Global Matching is straightforward to implement.  The Global + Content Matching version involves a series of detectors for objects, stuff, attributes, scene, and actions.  We use the same poselet-based action feature~\citep{Maji:2011:poselet-action}, which captures cues most relevant to our person-centric approach and utilizes the same \emph{ground truth} person bounding box used by our method.\footnote{We omit the object, stuff, and attribute detectors because we could not reproduce the implementation (hence the asterisk in the table).  In principle, any benefit from additional local content could also benefit us.}  For this application, we use our embedding method due to its advantage in handling large scale data such as COCO, and because the interaction fine-tuned feature can be directly applied to the retrieval-based sentence generation method (as opposed to the MDN, which produces probabilities for scales and positions, but no descriptor.

Table~\ref{tab:sentence-retrieval} shows the results. Our interaction embedding based non-parametric regression method consistently outperforms the baselines and~\citep{Berg:2011:im2text}.  The result confirms that a person-centric view of ``what to mention" is valuable.

We include an array of ablation studies to reveal the impact of the different features.  Without using CNN features, our non-parametric method (Ours-embedding w/o cnn) still outperforms the baselines~\citep{Berg:2011:im2text}.   The local Content Matching does not improve accuracy over Global Matching, and even detracts from it slightly.  We suspect this is due to weaknesses in poselets for this data, since the action variation is very high in COCO.  The authors also observed only a slight gain with Content Matching in their own results~\citep{Berg:2011:im2text}.  Note that like our method, the Content Matching method also has access to the correct person bounding box on a test image.  Our better results, therefore, cannot be attributed to having access to that information.

Our complete method (Ours-embedding w/all) provides higher accuracy than the baselines that also utilize CNN-based features (Global Matching$+$AlexNet fc7 and Global Matching$+$Places-CNN); the former uses CNN features from the person bounding box, while the latter uses CNN features from the whole image. To isolate the role of the interaction-guided CNN features, we show results when only those descriptors are used individually in conjunction with the non-parametric locally weighted regression.  As shown in Table~\ref{tab:sentence-retrieval}, our learned embedding features are helpful for the captioning task, guiding the system to focus on the interaction (Ours w/cnn-p $>$ Global$+$AlexNet, Ours w/cnn-s $>$ Global$+$Places-CNN).  Combining all the features provides the highest accuracy.  This indicates that features designed with domain knowledge (about gaze, pose, etc.) remain valuable in this setting, and can augment the automatically learned CNN features.

\begin{figure*}[t]
\centering
\includegraphics[width=1.0\textwidth]{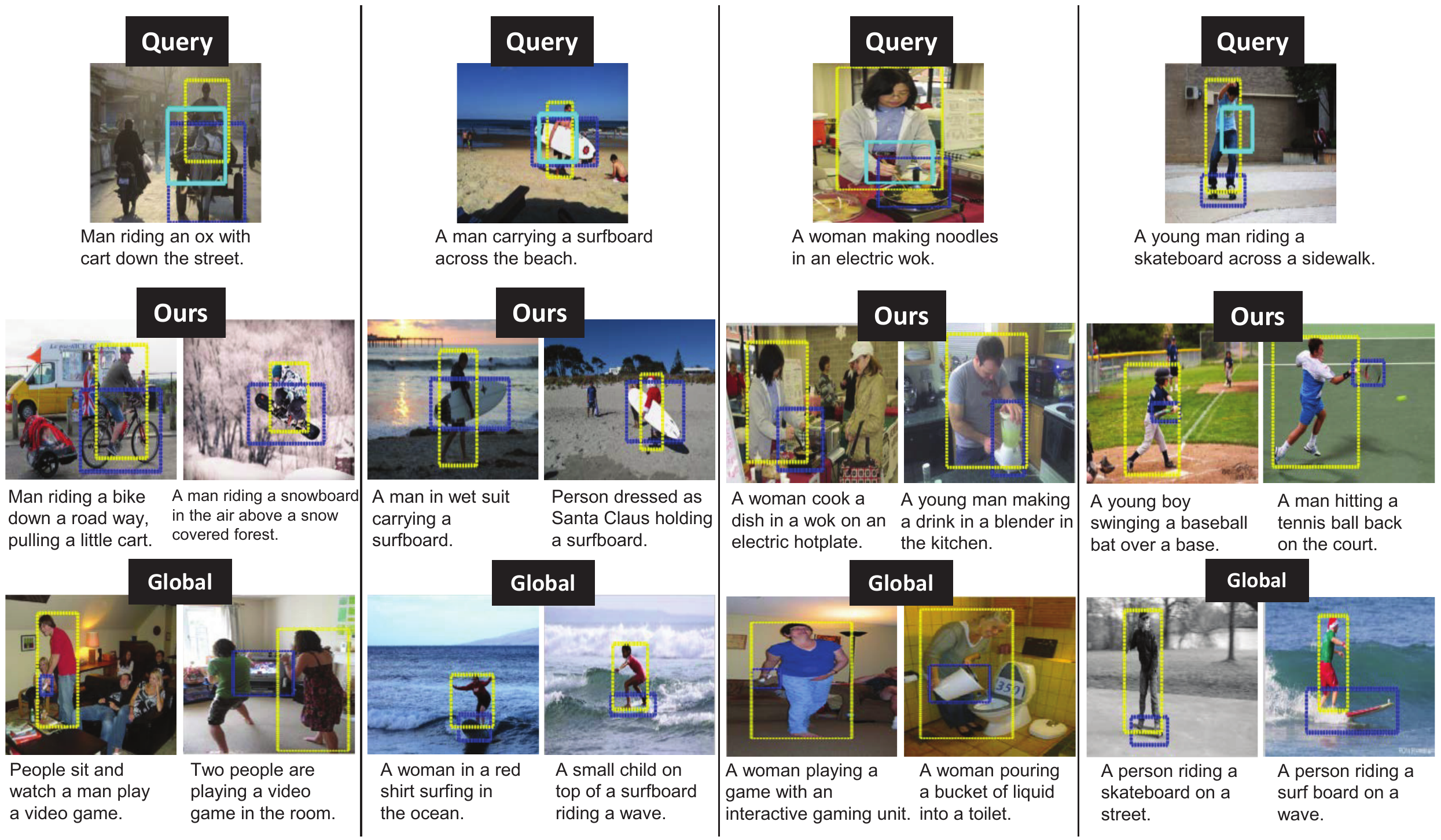}
   \caption[Example sentences generated by our method and baselines]{Example sentences generated by our method and the Global Matching method~\citep{Berg:2011:im2text}. Blue bbox: true interactee, cyan bbox: our prediction.  In the first three examples, ours is better because it correctly predicts the location of the interactee, and then uses the interactee's position and scale relative to the person to retrieve image examples with similar types of interaction.  In the last one, our method fails to predict the interactee correctly and thus retrieves poorly matched interactions. See text for details.}\label{fig:sentence}
\end{figure*}

Figure~\ref{fig:sentence} shows example sentences generated by our method, alongside those of the baselines.  We see how modeling person-centric cues of importance allows our method to find examples with similar interactions.  In contrast, the baselines based on global image matching find images focused on total scene similarity.  They often retrieve sentences describing similar overall scene contexts, but are unable to properly model the fine-grained interactions (e.g., in second column, riding vs.~carrying with a surfboard).  The fourth column shows a failure case by our method, where we mispredict the interactee (cyan box) and so retrieve people doing quite different interactions.

\makeatletter{}%
\section{Conclusions and Future Work}\label{sec:conclusion}

In this paper, we considered a new problem: how to predict where an interactee object will appear, given cues from the content of the image.  While plenty of work studies action-specific object interactions, predicting interactees in an action-independent manner is both challenging and practical for various applications.  The proposed method shows promising results to tackle this challenge.  We demonstrate its advantages over multiple informative baselines, including a state-of-the-art object saliency and importance metrics, and illustrate the utility of knowing where interactees are for contextual object detection, image retargeting, image description, and object importance ratings.  We also introduce a new 10,147-image dataset of interaction annotations for all person images in COCO.

In future work, we plan to extend the ideas to video, where an interactee will have a potentially more complex spatio-temporal relationship with its subject, yet dynamic cues may offer clearer evidence about the subject's attention.  We are also interested in exploring how more sophisticated language generation models could work in concert with our visual model of interactions.

\bibliographystyle{spbasic}      %
\bibliography{strings,ref}   %

\end{document}